\documentclass[runningheads,pagebackref,breaklinks,colorlinks]{llncs}
\usepackage{graphicx}
\usepackage{orcidlink}
\usepackage{tikz}
\usepackage{comment}
\usepackage{amsmath,amssymb} %
\usepackage{color}
\usepackage{multirow}
\usepackage[accsupp]{axessibility}  %
\usepackage{adjustbox}
\usepackage{booktabs}

\newif\ifdrafting
\draftingtrue %
\draftingfalse %
\ifdrafting
    \newcommand{\ds}[1]{{\color{green}[DS: #1]}}
    \newcommand{\cih}[1]{{\leavevmode\color[rgb]{0,0.4,0}[Charles: #1]}}
    \newcommand{\df}[1]{{\color{blue}[DF: #1]}}

\else
	\newcommand{\ds} [1] {}
	\newcommand{\cih} [1] {}
    \newcommand{\df}[1]{}	
	
\fi

\newcommand{\betweenfigure}{\vspace{0pt}}
\newcommand{\afterfigure}{\vspace{-0pt}}
\newcommand{\aftertable}{\vspace{-0pt}}

\newcommand{\fig}{Fig.} %

\newcommand{\vs}{\emph{v.s. }}
\newcommand{\etal}{\emph{et al.}}

\newcommand{\cf}{\emph{c.f.}}
\newcommand{\eg}{\emph{e.g.}}
 
\usepackage{pifont}%

\newcommand{\cmark}{\ding{51}}%
\newcommand{\xmark}{\ding{55}}%
\newcommand{\ignore}[1]{}

\begin{document}
\pagestyle{headings}
\mainmatter
\def\ECCVSubNumber{2206}  %

\title{Disentangling Architecture and Training for Optical Flow}

\titlerunning{Disentangling Architecture and Training for Optical Flow}
\authorrunning{Sun \etal } 

\author{ 
Deqing Sun\thanks{$\!\!$Equal technical contribution, $^{\dagger}$project lead.}$^{,\dagger}$ \orcidlink{0000-0003-0329-0456} %
 \qquad  %
Charles Herrmann$^{\star}$ \orcidlink{0000-0002-9576-9394}  \qquad 
Fitsum Reda \orcidlink{0000-0003-3072-4109}  \qquad  \\
 { Michael Rubinstein \orcidlink{0000-0002-3707-3807}  \qquad 
David J. Fleet\index{Fleet, David J.} \orcidlink{0000-0003-0734-7114}  \qquad 
William T. Freeman\index{Freeman, William T.}}\\
}
\institute{Google Research}
\maketitle

\begin{abstract}

How important are training details and datasets to recent optical flow architectures like RAFT?  And do they generalize?
To explore these questions, rather than develop a new architecture, we revisit three prominent architectures, PWC-Net, IRR-PWC and RAFT, with a common set of modern training techniques and datasets, and observe significant performance gains, demonstrating the importance and generality of these training details.
Our newly trained PWC-Net and IRR-PWC  show surprisingly large improvements, up to 30\% versus original published results on Sintel and KITTI 2015 benchmarks.
Our newly trained RAFT obtains an Fl-all score of 4.31\%  on  KITTI 2015 and an avg. rank of 1.7 for end-point error on Middlebury. %
{Our results demonstrate the benefits of separating the contributions of architectures, training techniques and datasets when analyzing performance gains of optical flow methods.}
Our source code is available at \url{https://autoflow-google.github.io}.

\end{abstract}

\section{Introduction}

The field of optical flow has witnessed rapid progress in recent years,
driven largely by deep learning.
FlowNet~\cite{Dosovitskiy:2015Flownet} first demonstrated the potential of deep learning for optical flow, while PWC-Net~\cite{sun2018pwc} was the first model to eclipse classical flow techniques. %
The widely-acclaimed RAFT model~\cite{teed2020raft} 
reduced error rates on common benchmarks by up to 30\% versus state-of-the-art baselines, outperforming PWC-Net by a wide margin.
RAFT quickly  became  the  predominant  architecture for optical flow~\cite{jiang2021learning,luo2022learning,mehl2021anisotropic,shi2022csflow,wan2020praflow_rvc,xiao2020learnable,xu2021flow1d,zhang2021separable} and related tasks~\cite{lipson2021raft,teed2021raft}. 

The success of RAFT has been attributed primarily to its novel architecture, including its multi-scale all-pairs cost volume, its recurrent update operator, %
and its up-sampling module. 
Meanwhile, other factors like training procedures and datasets have also evolved, and may play important roles.
In this work, we pose the question: %
How much do
training techniques of recent methods like RAFT contribute to their impressive performance?
And, importantly, can these training innovations %
similarly improve the performance of other architectures? %

\begin{figure}[t!]
\def\arraystretch{0.1}
\def\tabcolsep{2pt}
\begin{tabular}{l@{\hskip 3mm}r}
\noindent\parbox[t]{0.39\linewidth}{\includegraphics[width=\hsize]{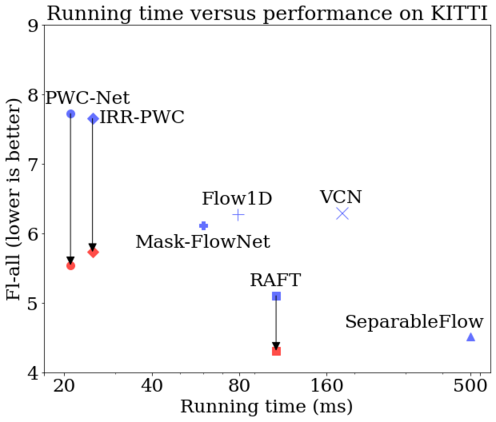}} &
\raisebox{22.5mm}{\parbox[t]{0.6\linewidth}{\centering \begin{tabular}{c@{\hskip 2mm}rcc}
{\scriptsize Inputs} & & {\scriptsize PWC-Net} & {\scriptsize RAFT} \\
\noindent\parbox[t]{0.3\linewidth}{\includegraphics[width=\hsize]{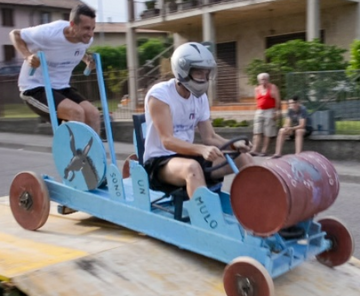}} & \raisebox{3mm}{\rotatebox{90}{\scriptsize Original}} &
\parbox[t]{0.3\linewidth}{\includegraphics[width=\hsize]{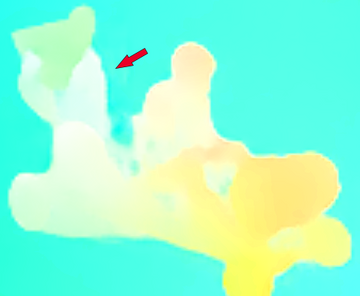}} &
\parbox[t]{0.3\linewidth}{\includegraphics[width=\hsize]{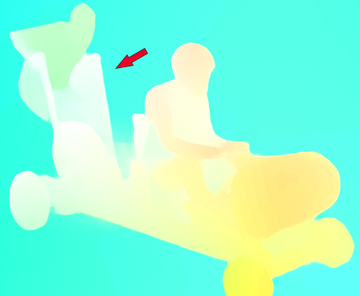}} \\
\noindent\parbox[t]{0.3\linewidth}{\includegraphics[width=\hsize]{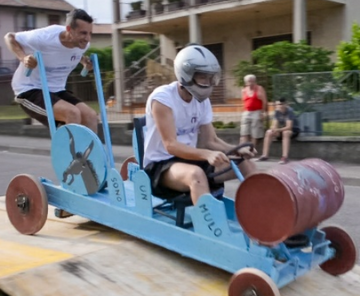}} & \raisebox{5.7mm}{\rotatebox{90}{{\scriptsize New (IT)}}} &
\parbox[t]{0.3\linewidth}{\includegraphics[width=\hsize]{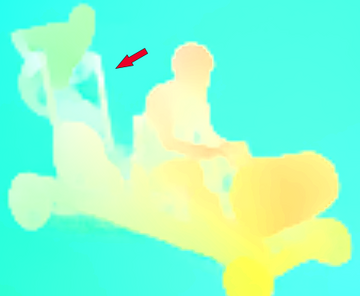}} &
\parbox[t]{0.3\linewidth}{\includegraphics[width=\hsize]{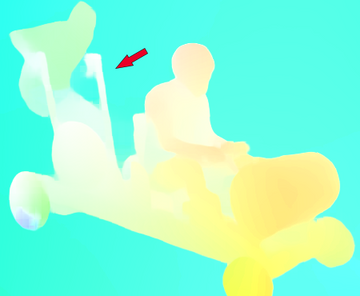}} \\
\end{tabular}}}\\
\end{tabular}
\caption{
Left: \textbf{Large improvements with newly trained PWC-Net, IRR-PWC and RAFT} (left: originally published results in {\color{blue}blue}; results of our newly trained models in {\color{red}red}).
Right: Visual comparison on a Davis sequence between the original~\cite{sun2019models} and our newly trained PWC-Net-it and RAFT-it, shows improved flow details, e.g. the hole between the cart and the person at the back. The newly trained PWC-Net-it recovers the hole between the cart and the front person better than RAFT. %
}
\label{fig:model:training:dataset}
\afterfigure
\end{figure}

We begin by revisiting the 2018 PWC-Net~\cite{sun2018pwc},
and investigate the impact of datasets %
and training techniques %
for both pre-training and fine-tuning.
We show that, even with such a relatively ``old'' model, by employing recent datasets and advances in training, {and without any changes to the originally proposed architecture}, one can obtain substantial performance gains, outperforming more recent models~\cite{xu2021flow1d,zhao2020maskflownet} and resolving finer-grained details of flow fields (see, e.g., Fig.~\ref{fig:model:training:dataset} and Table~\ref{tab:benchmark}). %
We further show that the same enhancements yield similar performance gains when applied to IRR-PWC, a prominent variant of PWC-Net that is closely related to RAFT. %
Indeed, these insights also yield an improved version of RAFT, which obtains competitive results on Sintel, KITTI, and VIPER while setting a new state of the art on Middlebury. We denote architectures trained with this new training by adding ``-it`` after the architecture name; for example, our newly trained RAFT will be abbreviated as RAFT-it.

We make the following contributions:
\begin{itemize}
    \item We show that newly trained PWC-Net (PWC-Net-it), using ingredients from recent training techniques (gradient clipping, OneCycle learning rate, and long training) and modern datasets  (AutoFlow), yields surprisingly competitive results on Sintel and KITTI benchmarks.
     \item These same techniques also deliver sizeable performance gains with two other prominent models, IRR-PWC and RAFT. Our newly trained RAFT (RAFT-it) is more accurate than all published optical flow methods on KITTI 2015. 
    \item We perform a thorough ablation study on pre-training and fine-tuning to understand which ingredients are key to these performance improvements and how they are manifesting. %
    \item The newly trained PWC-Net and IRR-PWC produce visually good results on 4K Davis input images, making them an appealing option for applications that require fast inference with low memory overhead.
\end{itemize}

\begin{table}[t]
\begin{center}
\small
\begin{tabular}{lcccc} 
Method & Sintel.clean & Sintel.final & KITTI & Running time \\ \hline
PWC-Net~\cite{sun2018pwc}  &  3.86 & 5.13  & 9.60\% & 30ms$^*$\\
PWC-Net+~\cite{sun2019models} &  3.45 & 4.60  & 7.72\% & 30ms$^*$ \\
PWC-Net-it (Ours)  & 2.31 & 3.69  & 5.54\% & \textbf{21ms}\\ \hline 
IRR-PWC~\cite{hur2019iterative} & 3.84 & 4.58 & 7.65\%  & 180ms$^*$\\ %
IRR-PWC-it (Ours) &2.19 & 3.55 & 5.73\% & \underline{25ms} \\  \hline
RAFT~\cite{teed2020raft} & {1.94} & {3.18}& {5.10}\% & 94ms$^*$  \\ %
RAFT-A~\cite{sun2021autoflow} & 2.01 & {3.14} &  {4.78}\% & 107ms\\
RAFT-it ~(Ours)  & \underline{1.55} & \underline{2.90}& \textbf{4.31\%} & 107ms\\ \hline 
HD$^3$\cite{yin2019hierarchical} & 4.79 & 4.67 & 6.55\% & 100ms$^*$  \\
VCN~\cite{yang2018segstereo} & 2.81 & 4.40 & 6.30\%  &180ms$^*$\\
Mask-FlowNet~\cite{zhao2020maskflownet} & 2.52	 & 4.14 &	6.11\%  & 60ms$^*$ \\
DICL~\cite{wang2020displacement} & 2.12 & 3.44 &  6.31\% & - \\
Flow1D~\cite{xu2021flow1d} & 2.24 & 3.81 & 6.27\% & 79ms$^*$ \\ 
RAFT+AOIR~\cite{mehl2021anisotropic} &1.85 & 3.17 & 5.07\% & 10$^4$ms$^*$\\
CSFlow~\cite{shi2022csflow} & 1.63 & 3.03 & 5.00\% & 200ms$^*$ \\
SeparableFlow~\cite{zhang2021separable} & \textbf{1.50} & \textbf{2.67} & \underline{4.51\%} & 250ms$^*$ \\ %
\end{tabular}
\caption{\textbf{Results of {2-frame} methods} on public benchmarks 
 (AEPE$\downarrow$ for Sintel and Fl-all$\downarrow$ for KITTI). \textbf{Bold} indicates the best number and \underline{underline} the second-best.
The running time is for $448 \!\times\! 1024$ resolution input ($^*$reported in paper); the differences will be larger for higher resolution (\cf~Table~\ref{tab:computational_complexity}). 
Newly trained PWC-Net, IRR-PWC and RAFT are substantially more accurate than their predecessors.
With improved training protocols,  PWC-Net-it and IRR-PWC-it are more accurate than some recent methods~\cite{xu2021flow1d,yang2019volumetric} on KITTI 2015 while being about $3\times$ faster in inference. %
}
\label{tab:benchmark}
\aftertable
\end{center}
\end{table}

\section{Previous Work}
\label{sec:prev}

\paragraph{Deep models for optical flow.}
FlowNet~\cite{Dosovitskiy:2015Flownet} was the first model to demonstrate the potential of deep learning for optical flow, and  inspired various new architectures.
FlowNet2~\cite{Ilg:2016:Flownet2} stacked basic models to improve model capacity and performance, while SpyNet~\cite{Ranjan:2016:SpyNet} used an image pyramid and warping to build a compact model. PWC-Net~\cite{sun2018pwc} used classical optical flow principles (e.g., \cite{Barron:1994:PO,sun2010secrets,szeliski2010computer}) to build an effective model, which has since seen widespread 
use~\cite{bao2019depth,djelouah2019neural,jiang2018super,kim2019deep,ranjan2019competitive,stroud2020d3d,zhao2019sound,zhao2020suppress}. %
The concurrent LiteFlowNet~\cite{Hui_2018_CVPR} used similar ideas to build a lightweight network. TVNet~\cite{Fan_2018_CVPR} took a different approach with classical flow principles by unrolling the optimization iterations of the TV-L1 method~\cite{zach2007duality}. 

Many architectures have used a pyramid structure. IRR-PWC~\cite{hur2019iterative} introduced iterative refinement, reusing the same flow decoder module at different pyramidal levels. 
VCN~\cite{yang2019volumetric} used a 4D cost volume that is easily adapted to stereo and optical flow. HD$^3$~\cite{yin2019hierarchical} modeled flow uncertainty hierarchically.
MaskFlowNet~\cite{zhao2020maskflownet} jointly modeled occlusion and optical flow.  Improvements brought by each model over the previous SOTA was often within 5\% on Sintel (\cf, Table~\ref{tab:benchmark}).

A recent, notable architecture,  RAFT~\cite{teed2020raft}, 
built a full cost volume and performs recurrent refinements at a single resolution. RAFT achieved a significant improvement over previous models on Sintel and KITTI benchmarks, and became a starting point for numerous new variants~\cite{jiang2021learning,luo2022learning,mehl2021anisotropic,shi2022csflow,wan2020praflow_rvc,xiao2020learnable,xu2021flow1d,zhang2021separable}. %
To reduce the memory cost of the all-pairs cost volume, Flow1D used 1D self-attention with 1D search, with minimal performance drop while enabling application to 4K video inputs~\cite{xu2021flow1d}. SeparableFlow used a  non-local aggregation module for cost aggregation, yielding substantial performance gains~\cite{zhang2021separable}.

Recent research on optical flow has focused on architectural innovations.
Nevertheless, most new optical flow papers combine new architectures with changes in training procedures and datasets. 
As such, it can be hard to identify which  factors are responsible for the performance gains.
In this paper, we take a different approach, instead we examine the effects of different ingredients of modern training techniques and datasets, but with established architectures.
The results and findings are surprising.
Our newly trained PWC-Net and IRR-PWC are more accurate than Flow1D while being almost $3\mathord\times$ faster in inference, and our newly trained RAFT is more accurate than all published optical flow methods on KITTI 2015 while being more than $2\mathord\times$ faster in inference than the previous best SeparableFlow. 

\paragraph{Datasets for optical flow.}
For pre-training the predominant dataset is FlyingChairs~\cite{Dosovitskiy:2015Flownet}. Ilg \etal~\cite{Ilg:2016:Flownet2} introduced a dataset schedule that uses FlyingChairs and FlyingThings3D~\cite{Mayer:2016:Large} sequentially.
This remains a standard way to pre-train models. 
Sun \etal~\cite{sun2021autoflow} proposed a new dataset, AutoFlow, which learns rendering hyperparameters and shows moderate improvements over the FlyingChairs and FlyingThings3D in pre-training PWC-Net and RAFT. For fine-tuning, the limited training data from Sintel and KITTI are often combined with additional datasets, such as HD1K~\cite{kondermann2016hci} and VIPER~\cite{richter2017playing}, to improve generalization. 
In this paper, we show that PWC-Net and its variant, IRR-PWC, obtain competitive results when pre-trained on AutoFlow and fine-tuned using recent techniques.

\paragraph{Training techniques for optical flow.}
While different papers tend to adopt slightly different training techniques and implementation details, some have examined the impact of recent training techniques on older architectures.
Ilg \etal~\cite{Ilg:2016:Flownet2} found that using dataset scheduling can improve the pre-training results of FlowNetS and FlowNetC.
Sun \etal~\cite{sun2019models} obtained better fine-tuning results with FlowNetS and FlowNetC on Sintel by using improved data augmentation and learning rate disruption;
they also improved on the initial PWC-Net \cite{sun2018pwc} by using additional datasets.
Sun \etal~\cite{sun2021autoflow} reported better pre-training results for PWC-Net but did not investigate fine-tuning. Here, with PWC-Net, IRR-PWC and RAFT, we show significantly better fine-tuning results.

\paragraph{Self-supervised learning for optical flow.}
Significant progress has been achieved with self-supervised learning for optical flow~\cite{jonschkowski2020matters,liu2019selflow,meister2018unflow,liu2020learning,stone2021smurf,yu2016back}, focusing more on the loss than model architecture. UFlow~\cite{jonschkowski2020matters} systematically studied a set of key components for self-supervised optical flow, including both model elements and training techniques. 
Their study used PWC-Net as the main backbone. Here we focus on training techniques and datasets,
systematically studying three prominent models to identify factors that generalize across models. FOAL introduces a meta learning approach for online adaptation~\cite{yu2020foal}.

\paragraph{Similar study on other vision tasks.}
The field of classification has also started to more closely examine whether performance improvements in recent papers come from the model architecture or training details. Both~\cite{he2019bag} and~\cite{wightman2021resnet} examined modern training techniques on ResNet-50~\cite{he2016deep} and observed  significant performance improvements on ImageNet~\cite{deng2009imagenet}, improving top-1 precision from 76.2 in 2015, to 79.3 in 2018, and finally to 80.4 in 2021. 
These gains have come solely from improved training details, namely, from augmentations, optimizers, learning rate schedules, and regularization. The introduction of vision transformers (ViT)~\cite{dosovitskiy2020image} also led to a series of papers~\cite{steiner2021train,touvron2021training} on improved training strategies, substantially improving performance from the initial accuracy of 76.5 up to 81.8.

Other recent papers took a related but slightly different direction, simultaneously modernizing both the training details and architectural elements but cleanly ablating and analyzing the improvements. Bello~\etal~\cite{bello2021revisiting} included an improved training procedure as well as exploration of squeeze-and-excite and different layer changes.  Liu~\etal~\cite{liu2022convnet} used recent training details and iteratively improves ResNet with modern network design elements, improving the accuracy from 76.2 to 82.0, which is competitive with similarly sized state-of-the-art models. 
While these papers mainly studied a single model and often involved modifying the backbone, we investigate three different models to understand key factors that apply to different models, and the trade-offs between models.

\section{Approach and Results}
\label{sec:approach}

{
Our goal is to understand which innovations in training techniques, principally from RAFT, play a major role in the impressive performance of modern optical flow methods, and to what extent they generalize well to different architectures.
To this end, we decouple the contributions of architecture, training techniques, and dataset, and perform comparisons by changing one variable at a time. 
More specifically, we revisit PWC-Net, IRR-PWC and RAFT with the recently improved training techniques and datasets. We perform ablations on various factors including pre-training, fine-tuning, training duration, memory requirements and inference speed.
}

\subsection{Models Evaluated}
The first model we evaluate is PWC-Net, the design of which was inspired by three classical optical flow principles, namely pyramids, warping, and cost volumes. These inductive biases make the network effective, efficient, and compact compared to prior work. IRR-PWC~\cite{hur2019iterative} introduces iterative refinement and shares the optical flow estimation network weights among different pyramid levels. The number of iterative refinement steps for IRR-PWC is the number of pyramid levels.
RAFT is closely related to IRR but enables an arbitrarily large number of refinement iterations. It has several novel network design elements, such as the recurrent refinement unit and convex upsampling module. Notably, RAFT eschews the pyramidal refinement structure, 
instead using an all-pairs cost volume at a single resolution.

\paragraph{Memory usage.}
For an $H\!\times\!W$ input image, the memory cost for constructing the cost volume in RAFT is $\mathcal{O}( (HW)^2 D)$, where $D$ is the number of feature channels (constant, typically 256 for RAFT and $\leq 192$ for PWC-Net and IRR-PWC). To reduce the memory cost for high-resolution inputs, Flow1D constructs a 1D cost volume with cost of $\mathcal{O}( HW(H\!+\!W)D)$. 
By comparison, the memory needed for the cost volume in PWC-Net and IRR-PWC is $\mathcal{O}(HWD(2d\!+\!1)^2)$, where the constant $d$ is the search radius at each pyramid level (default $4$). Note that $(2d\!+\!1)^2 \ll H\!+\!W \ll HW$ for high-resolution inputs; this is particularly important for 4K videos, which are becoming increasingly popular. We  empirically compare memory usage at different resolutions in Table~\ref{tab:computational_complexity}.

\subsection{Pre-training }

\paragraph{Typical training recipes.}
A typical training pipeline trains models first on the FlyingChairs dataset, followed by fine-tuning on the FlyingThings3D dataset, and then further fine-tuning using a mixture of datasets, including small amount of training data for the Sintel and KITTI benchmarks. 

Since the introduction of PWC-Net in 2018,  new training techniques and datasets have been proposed. As shown in~\cite{sun2019models}, better training techniques and new datasets improve the pre-training performance of PWC-Net. We investigate how PWC-Net and IRR-PWC performs with the same pre-training procedure, and whether the procedure can be further improved.  

Table~\ref{tab:pwcnet:pre} summaries the results of pre-training PWC-Net, IRR-PWC and RAFT using different datasets and techniques. 
(To save space, we omit some results for PWC-Net and RAFT and refer readers to~\cite{sun2021autoflow}.)
We further perform an ablation study on several key design choices using PWC-Net, shown in Table~\ref{tab:pwcnet:pre:ablation}. %
To reduce the effects of random initialization, we independently 
train the model six times, and report the results of the best run. While the original IRR-PWC computes bidirectional optical flow and jointly reasons about occlusion, we test a lightweight implementation without these elements~\cite{hur2019iterative}.%

\begin{figure*}[ht!]
    \centering
    \small
    \newcommand{\Figwidth}{0.32\linewidth}
    \begin{tabular}{ccc}
    \betweenfigure    First frame & \betweenfigure PWC-Net (FlyingChairs) & \betweenfigure PWC-Net (AutoFlow) \\
    \betweenfigure    
    \includegraphics[trim={0 0 0 0},clip,width=\Figwidth]{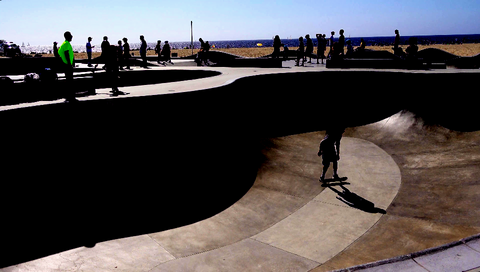} &
    \betweenfigure
    \includegraphics[trim={0 0 0 0},clip,width=\Figwidth]{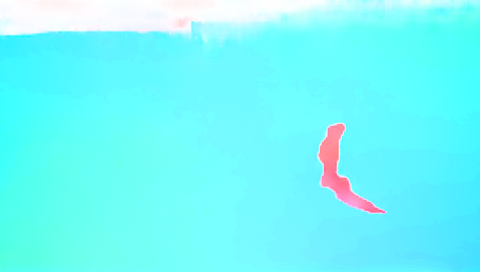} &
    \betweenfigure
    \includegraphics[trim={0 0 0 0},clip,width=\Figwidth]{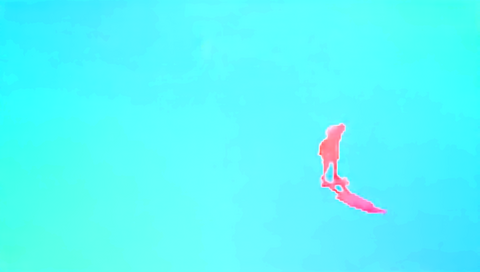} \\
    \betweenfigure
    \includegraphics[trim={0 0 0 0},clip,width=\Figwidth]{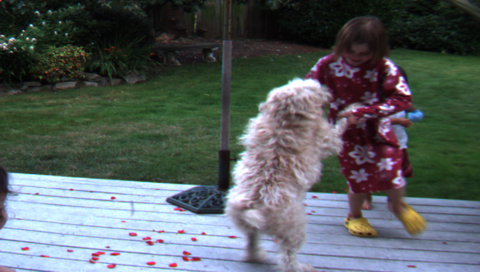} &
    \betweenfigure
    \includegraphics[trim={0 0 0 0},clip,width=\Figwidth]{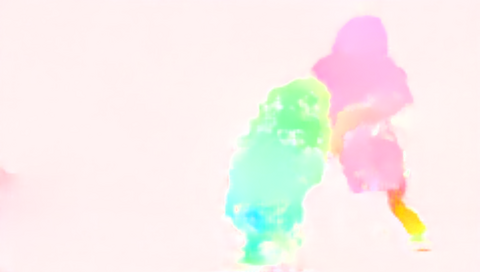} &
    \betweenfigure
    \includegraphics[trim={0 0 0 0},clip,width=\Figwidth]{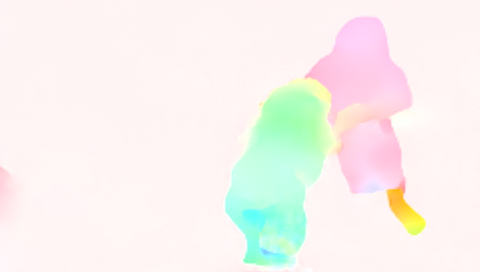} \\

    \end{tabular}
	\caption{Visual results of PWC-Net pre-trained using FlyingChairs and AutoFlow on Davis 
	and Middlebury input images. PWC-Net trained using AutoFlow recovers fine details between the legs (top) and coherent motion for 
	the girl and the dog (bottom).  }
    \label{fig:pretraining_chairs_vs_raft}
\end{figure*}

\paragraph{Pre-training datasets.} Pre-training using AutoFlow results in significantly better results than FlyingChairs for PWC-Net, IRR-PWC and RAFT. %
Figure~\ref{fig:pretraining_chairs_vs_raft} visually compares the results by two PWC-Net models on Davis~\cite{Perazzi2016}, %
and Middlebury~\cite{Baker:2011:DEO} sequences. PWC-Net trained on AutoFlow better recovers fine motion details (top) and produces coherent motion for the foreground objects (bottom).

\paragraph{Gradient clipping.} Gradient clipping is a heuristic to avoid cliff structures for recurrent neural networks~\cite{goodfellow2016deep}. The update operator of RAFT uses a GRU block that is  similar to the LSTM block. Thus, RAFT training uses gradient clipping to avoid exploding gradients. Gradient clipping also improves the performance of PWC-Net and IRR-PWC substantially and results in more stable training. 
Removing gradient clipping from RAFT results in moderate performance degradation. 
We perform an ablation study on the threshold of gradient clipping and find that the training is robust to this parameter  (Table~\ref{tab:pwcnet:pre:ablation}).

\paragraph{Learning rate schedule.} Before RAFT, nearly all optical flow models have been trained using a piecewise learning rate, with optional learning rate disruption. RAFT uses a OneCycle learning rate schedule, which starts from a small learning rate, linearly increases to the peak learning rate, and then linearly decreases to the starting learning rate. Using the OneCycle learning rate improves the  performance of all three models (Table~\ref{tab:pwcnet:pre}). Moving the position of the peak toward the origin slightly improves the performance (Table~\ref{tab:pwcnet:pre:ablation}). Note that, for other published models, those that use gradient clipping and the OneCycle learning rate, \eg, Flow1D and SeparableFlow, are generally better than those that do not, \eg, VCN and MaskFlowNet. It would be interesting, though outside the scope of this paper, to investigate the performance of VCN and MaskFlowNet with recent techniques and datasets.

\paragraph{Training iterations.} %
PWC-Net and IRR-PWC need large numbers of training iterations.
At the same number of training iterations, IRR-PWC is consistently more accurate than PWC-Net. This is encouraging because we can perform an ablation study using fewer iterations and then use the best setup to train the model using more iterations. 
One appealing feature of RAFT is its fast convergence, but we find that using more training iterations also improves RAFT. 
Note that 3.2M iterations for RAFT takes about 11 days  while 6.2M iterations take PWC and IRR-PWC about 6 days to finish (using 6 P100 GPUs). It is interesting that all three models show no sign of over-fitting after so many iterations.

\paragraph{Other training details.}
We further test the effect of weight decay, random erasing and vertical flipping. As shown in Table~\ref{tab:pwcnet:pre:ablation}, the training is  robust to the hyperparameter settings for the weight decay, random erasing and vertical flipping. %

\newcommand{\pwcsintelclean}{2.17}
\newcommand{\pwcsintelfinal}{2.91}
\newcommand{\pwckitti}{5.76}

\begin{table}[ht]
\begin{center}
\small
\begin{tabular}{ ccccc|cccc } 
\multirow{2}{*}{Model} & \multirow{2}{*}{Dataset} & \multirow{2}{*}{GC} & \multirow{2}{*}{LR}  & \multirow{2}{*}{Iters} & \multicolumn{2}{c}{Sintel} & \multicolumn{2}{c}{KITTI} \\ 
& & & &   & clean & final & F-all  & AEPE \\ \hline
PWC-Net& FlyingChairs & \xmark & Piecewise &1.2M & 3.89	 & 4.79 &	42.81\%	& 13.59 \\ %
-& - &- & - &3.2M &2.99&	4.21	&38.49\%	&10.7 \\ %
-& AutoFlow & \cmark  & OneCycle&  1.2M &2.43 &	3.05	&18.74\% &	6.41\\
-& - & -  & - & 3.2M & {2.17} &	{2.91} & 17.25\%  & {5.76} \\
-& - & -  & - & 6.2M & 2.10 &	2.81& 16.29\%  & 5.55   \\ \hline
IRR-PWC & FlyingChairs & \xmark & Piecewise & 1.2M &4.3	&5.09	&44.06\%	&15.5\\
-& AutoFlow & - & - & -  &3.01	&4.11	&26.95\%	&9.01 \\ 
-& - & \cmark & - & - &2.42	&3.29	&18.31\%	&6.31  \\
- &  - &  - & OneCycle& - &2.24&	2.93&	17.87\%&	6.02 \\ 
- &  - & -  & -& 3.2M &2.06	&2.85&	15.55\%&	5.14 \\ 
- &  - & -  & -& 6.2M &1.93	&2.76 & 15.20\%		&5.05 \\  \hline
RAFT & FlyingChairs & \xmark & Piecewise    & 0.2M  &2.64	&4.04	&32.52\%	&10.01 \\
- & AutoFlow & -  & -  &- &2.57	&3.36	&19.92\%	&5.96			\\
-& - &  \cmark  & -   & - &2.44	&3.20	&17.95\%	&5.49\\
-& - &  - & OneCycle  &  - & 2.08 &	2.75	& 15.32\% &	4.66\\
-& - & -  & - & 0.8M &1.95 &  2.57 & 13.82\% & 4.23 \\
-& - & -  & - & 3.2M &1.74	&2.41& 13.41\%	&4.18  \\ \hline	%
VCN & C+T  & \xmark & Piecewise &   0.22M &2.21 &3.62 & 25.10\% & 8.36 \\
MaskFlowNet& - & - & - & 1.7M & 2.25 & 3.61 & 23.14\% & - \\
Flow1D& - & \cmark  & OneCycle  &0.2M & 1.98 & 3.27 &  22.95\% & 6.69 \\
SeparableFlow& - & - &-  &  0.2M & 1.30 & 2.59 & 15.90\% & 4.60 \\
\end{tabular}
\caption{ \textbf{Pre-training} results for PWC-Net, IRR-PWC,  RAFT and some recent methods. The metric for Sintel is average end-point error (AEPE) and F-all is the percentage of outliers averaged over all ground truth pixels.  Lower is better for both AEPE and F-all. ``-'' means the same as the row above. C+T stands for the FlyingChairs and FlyingThings3D dataset schedule. 
Gradient clipping (GC), OneCycle learning rate, AutoFlow and longer training  improve all three models consistently. 
}
\label{tab:pwcnet:pre}
\end{center}
\aftertable
\end{table}

\begin{table}[ht]
\begin{center}
\small
\begin{tabular}{ llcc|cc } 
\multirow{2}{*}{Experiment} & \multirow{2}{*}{Parameter}& \multicolumn{2}{c}{Sintel} & \multicolumn{2}{c}{KITTI} \\ 
& &  clean & final& F-all  & AEPE \\ \hline
\multirow{3}{*}{Gradient clipping threshold} & 0.5 &2.37	&3.12	&18.46\%	&6.14 \\
& \underline{1.0} &2.43 &	3.05	&18.74\% &	6.41  \\
& 2.0 &2.60	&3.31	&21.25\%	&7.73 \\  \hline
\multirow{3}{*}{Peak of OneCycle LR} & 0.1 &2.38	&3.04	&17.35\%	&5.77 \\
& \underline{0.2 }&2.43 &	3.05	&18.74\% &	6.41  \\
& 0.3 &2.35	&3.08	&19.39\%	&6.66   \\ \hline
\multirow{3}{*}{Weight decay} & \underline{0} &2.43 &	3.05	&18.74\% &	6.41  \\
& 1e-8 & 2.31	& 3.09	& 18.07\%	& 6.14	\\
& 1e-7 & 2.46	& 3.17	&18.10\% &	6.17\\  \hline
\multirow{2}{*}{Vertical flip probability} &  \underline{0} &2.43 &	3.05	&18.74\% &	6.41  \\
&  {0.1} &2.38	&3.08	&18.64\%	&6.14 \\  \hline
\multirow{2}{*}{Random erasing probability} &  \underline{0} &2.43 &	3.05	&18.74\% &	6.41 \\
& 0.5 &2.46	&3.13	&17.39\%	&5.78\\
\end{tabular}
\caption{ \textbf{More ablation studies} on pre-training PWC-Net using 1.2M training steps. Default settings are underlined. Pre-training is robust to moderate variations on the parameters settings for these training details. 
}
\label{tab:pwcnet:pre:ablation}
\aftertable
\end{center}
\end{table}

\paragraph{Recipes for Pre-training.}
Using AutoFlow, gradient clipping, the OneCycle learning rate and long training consistently improves the pre-training results for PWC-Net, IRR-PWC and RAFT. It is feasible to use short training to evaluate design choices and then use longer training times for the best performance.

\subsection{Fine-tuning}
\begin{figure*}[ht!]
    \begin{center}
    \small
        \newcommand{\figwidth}{0.035\linewidth}
        \newcommand{\Figwidth}{0.46\linewidth}
        \newcommand{\Figheight}{0.20\linewidth}
        \begin{tabular}{ccc}
        \includegraphics[height = \Figheight, width=\Figwidth]{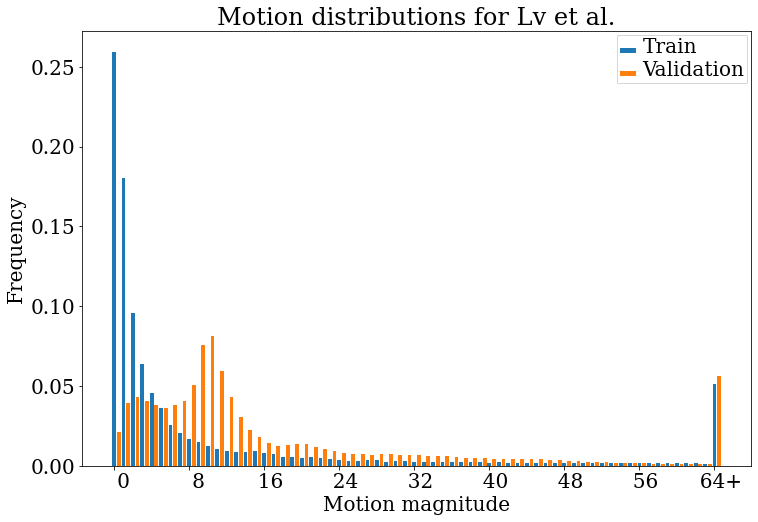}  & &
        \includegraphics[height = \Figheight, width=\Figwidth]{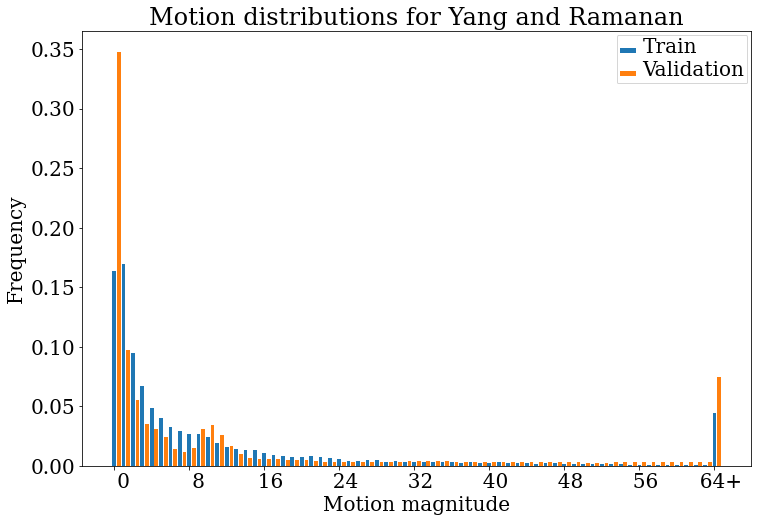}   \\    
        \end{tabular}
		\caption{Motion distributions for the Lv \etal~\cite{Lv18eccv} (left) and Yang and Ramanan~\cite{yang2019volumetric} (right) training/validation splits. There is a mismatch between training and validation distributions for the Lv split, making it suitable for out-of-distribution fine-tuning test, while the other split is more suitable for in-distribution test.
		}	
        \label{fig:trainval:split}
        \afterfigure
    \end{center}
\end{figure*}

To analyze fine-tuning, we use the training/validation split for Sintel proposed in Lv \etal~\cite{Lv18eccv}, where the sets have different motion distributions  (\fig~\ref{fig:trainval:split}), and the training/validation split for KITTI proposed in Yang and Ramanan~\cite{yang2019volumetric}. We  follow~\cite{sun2021autoflow} and use five datasets, Sintel~\cite{Butler:ECCV:2012} (0.4), KITTI~\cite{Geiger:2012:We} (0.2), VIPER~\cite{Richter_2017} (0.2), HD1K~\cite{kondermann2016hci} (0.08), and FlyingThings3D~\cite{Mayer:2016:Large} (0.12), where the number indicates the sampling probability. We perform an ablation study on PWC-Net, and then apply the selected training protocol to IRR-PWC and RAFT. %

\begin{table}[ht]
\def\tabcolsep{1pt}
\begin{center}
\footnotesize
\begin{tabular}{ lccc|cccccccc } 
& & & & \multicolumn{4}{c}{Sintel} & \multicolumn{4}{c}{KITTI 2015} \\
Model & Data & Init & Ft  & \multicolumn{2}{c}{Training} & \multicolumn{2}{c}{Validation} & \multicolumn{2}{c}{Training} & \multicolumn{2}{c}{Validation} \\
& &  &  & clean & final    & clean & final &  F-all & AEPE &  F-all & AEPE \\ \hline
PWC-Net &SKHTV & 1.2M & 1.2M & (1.04)	&(1.45)	&3.58	&3.88	&(5.58\%)	&(1.44)	&6.23\%	&1.92 \\
- &- & 3.2M &- &(1.05)	&(1.55)	&2.95	&3.61 &(5.44\%)	&(1.40)	&6.13\%	&1.80   \\
- & - & 6.2M & - &(0.97)	&(1.42)	&3.09	&3.65 &(4.99\%)	&(1.31)	&5.61\%	&1.62 \\ 
No GC &- &- &-  &(1.82)&	(2.43)&	4.51	&4.98	&(11.77\%) & (3.06)&	11.87\%	&3.79 \\
Piecewise &- &- & - &(1.08)&	(1.62)&	3.32	&3.77	&(5.49\%) &(1.42) &5.90\%	&1.78 \\ \hline
PWC-Net & SKHTV & 6.2M & 0M & 1.78	& 2.55	& 3.33	& 3.83 &16.50\%&	5.58	&15.45\%&	5.44\\
- & - &- & 6.2M  &(0.74)	&(1.08)	&2.79	&3.52	&(3.96\%)	&(1.08)	&4.76\%	&1.52 \\
-&+A &- &-  & (0.80)	& (1.19)	 &2.76 &	3.25  &  (4.10\%)	& (1.12)	&	4.89\%	&1.57 \\ \hline
IRR-PWC  &SKHTV &6.2M &  0M &1.58	&2.49	&3.27	&3.79 &15.4\% &	5.05	&14.3\%&	5.02\\
- & - & - & 6.2M &(0.98)	&(1.47)	&2.85	&3.50	&(4.52\%)	&(1.21)	&5.37\%	&1.59\\
- & +A &- & - &(1.01)	&(1.49)	&2.64	&3.28 & (4.86\%)	& (1.29)	&5.39\%	&1.56 \\ \hline
RAFT  &SKHTV &  3.2M &  0M& 1.40	&2.31	&2.88	&3.38  &13.57\%&	4.19	&12.74\%&	4.13\\ %
- & - & - & 1.2M &(0.66)	&(1.14)	&1.96	&2.81	& (3.55\%)	& (1.04)	&3.96\%	&1.41 \\
- & +A & - & - &(0.74)	&(1.15)	&2.00	&2.76	&(3.86\%)	&(1.09)	&4.08\%	&1.39 \\
\end{tabular}
\caption{ \textbf{Ablation study } on fine-tuning on Sintel and KITTI using the training/validation split for Sintel from ~\cite{Lv18eccv} and for KITTI from~\cite{yang2019volumetric}. GC stands for gradient clipping and () indicates training errors. 0M for fine-tuning means that no fine-tuning has been done (initialization).
S,K,H,T,V and A denote Sintel, KITTI, FlyingThings3D, HD1K, VIPER and AutoFlow datasets, respectively. Better initialization, more training steps and adding AutoFlow improve the performance. 
}
\label{tab:ft:sintel:lv:ablation}
\aftertable
\end{center}
\end{table}
\paragraph{Training techniques.}
Table~\ref{tab:ft:sintel:lv:ablation} summarizes the results of the ablation study on PWC-Net. 
Better initialization tends to lead to better fine-tuning results, especially on the KITTI dataset. For the same initialization, longer training yields more accurate results on the held-out validation set.

Removing gradient clipping results in a significant performance drop on the validation sets, and switching from the OneCycle to the piecewise learning rate results in moderate performance degradation too. 
We further further experiment with adding the AutoFlow data to the fine-tuning process, and observe improvements for both PWC-Net and IRR-PWC on the Sintel validation set, and a small drop in performance on the KITTI validation set. 
Adding AutoFlow yields just a small improvement for RAFT on Sintel (we discuss this result again below with the in-distribution fine-tuning experiment).

\paragraph{Model comparison.}
Among the three models, RAFT has the best accuracy on the validation set. The initialization of RAFT is almost as accurate as the fine-tuned PWC-Net on the Sintel.final validation set using the training/validation split~\cite{Lv18eccv}. While IRR-PWC has higher training errors on Sintel than PWC-Net, the validation errors of
the two models are similar. IRR-PWC has slightly worse performance on the KITTI validation set than PWC-Net. 

\paragraph{In-distribution fine-tuning.}
The training and validation subsets for Sintel proposed by Lv \etal~\cite{Lv18eccv} have different motion distributions; the validation set has more middle-to-large range motion, as shown in \fig~\ref{fig:trainval:split}. To examine the performance of fine-tuning when the training and validation sets have similar distributions, we perform fine-tuning experiments using another split by~\cite{yang2019volumetric}. As summarized in Table~\ref{tab:ft:sintel:yang}, PWC-Net has lower errors than RAFT on the Sintel validation set. 
As shown in \fig~\ref{fig:trainval:split}, both the training and validation sets by~\cite{yang2019volumetric} concentrate on small motions, suggesting that RAFT is good at generalization to out-of-distribution large motion for the Lv \etal{} split. This generalization behavior likely explains why adding AutoFlow~\cite{sun2021autoflow} does not significantly help RAFT in the experiment above.
The result also suggests that PWC-Net may be a good option for applications dealing with small motions, \eg, the hole between the cart and the man in the front in \fig~\ref{fig:model:training:dataset}. 

\begin{table}[ht]
\begin{center}
\small
\begin{tabular}{ lcccccccc } 
& \multicolumn{4}{c}{Sintel} & \multicolumn{4}{c}{KITTI 2015} \\
 & \multicolumn{2}{c}{Training} & \multicolumn{2}{c}{Validation} & \multicolumn{2}{c}{Training} & \multicolumn{2}{c}{Validation}\\ 
& clean & final    & clean & final &  F-all & AEPE &  F-all & AEPE \\ \hline
PWC-Net &  2.06	& 2.67	& 2.24	& 3.23 &16.50\%&	5.58	&15.45\%&	5.44\\
PWC-Net-ft  &(1.30)	&(1.67)	&1.18	&1.74	&(4.21\%)	&(1.14)	&5.10\%	&1.51 \\ %
\hline
IRR-PWC &1.87	&2.53	&2.09	&3.44  &15.4\% &	5.05	&14.3\%&	5.02 \\
IRR-PWC-ft &(1.34)	&(1.88)	&1.55	&2.31	&(4.94\%)	&(1.29)	&5.42\%	&1.65 \\ %
\hline
RAFT &1.74	&2.24	&1.74	&2.91  &13.57\%&	4.19	&12.74\%&	4.13\\ 
RAFT-ft &(1.14)	&(1.70)	&1.37	&2.14	&(5.06\%)	&(1.61)	&5.01\%	&1.40 \\ %
\end{tabular}
\caption{ \textbf{In-distribution} fine-tuning using the training/validation split~\cite{yang2019volumetric} for Sintel. 
The training and validation sets share similar motion distributions (\cf \fig~\ref{fig:trainval:split}).
}
\label{tab:ft:sintel:yang}
\aftertable
\end{center}
\end{table}

\paragraph{Recipes for Fine-tuning.} Using better initialization and long training times helps fine-tuning. Both gradient clipping and the OneCycle learning rate help fine-tuning. Adding AutoFlow may help
with generalization of the models.

\begin{figure}[ht!]
    \centering
    \includegraphics[trim={0 0 0 0},clip,width=0.95\linewidth]{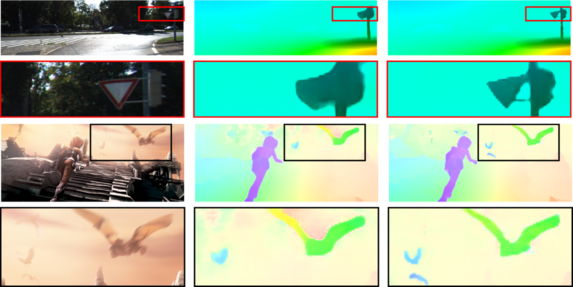}
	\caption{\textbf{Representative visual results} on KITTI and Sintel test sets by the original~\cite{sun2019models} and our newly trained PWC-Net (both fine-tuned).  Our newly trained PWC-Net can better recover fine details, \eg, the traffic sign (top) and  the small birds and the dragon's right wing (green is correct, bottom).}	
    \label{fig:pwc}
    \afterfigure
\end{figure}

\subsection{Benchmark Results}

We next apply the fine-tuning protocols above, with the full training sets from KITTI and Sintel, and then test the fine-tuned models on the public test sets. Table~\ref{tab:benchmark} summarizes the {2-frame} results of previously published PWC-Net, IRR-PWC, and RAFT, our newly trained models, and several recent methods.

\paragraph{MPI Sintel.}
Our newly trained PWC-Net-it and IRR-PWC-it are substantially better than the respective, published models, with up to a 1 pixel reduction in average end-point error (AEPE) on the Sintel benchmark. As shown in \fig~\ref{fig:pwc}, PWC-Net-it can much better recover fine motion details than the published one~\cite{sun2019models}. PWC-Net-it and IRR-PWC-it are even more accurate than some recent models~\cite{yang2019volumetric,zhao2020maskflownet,xu2021flow1d} on the more challenging final pass, while being about $3\times$ faster during inference. 

\begin{figure*}[t!]
  \centering
  \small
  \setkeys{Gin}{width=0.2\linewidth}
  \resizebox{\textwidth}{!}{\begin{tabular}{ccccc}
  First frame & Ground truth & RAFT & SeparableFlow & RAFT-it (ours) \\
  \includegraphics{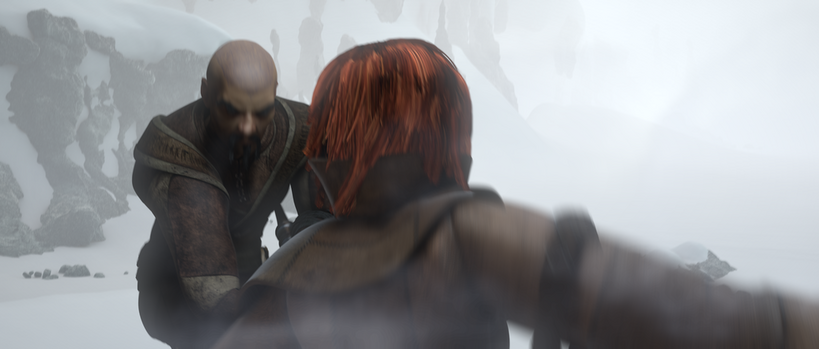}&
  \includegraphics{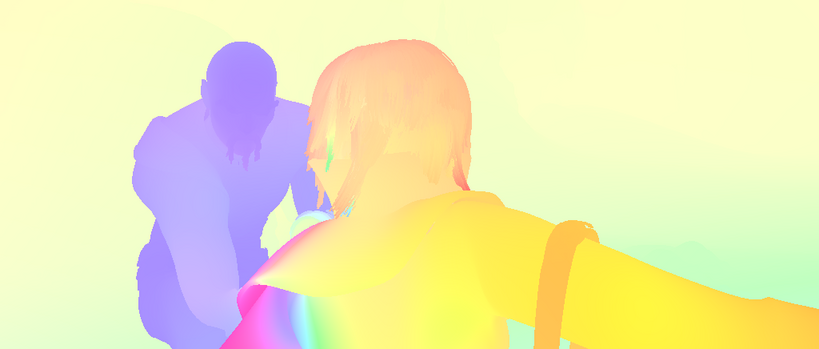}&
  \includegraphics{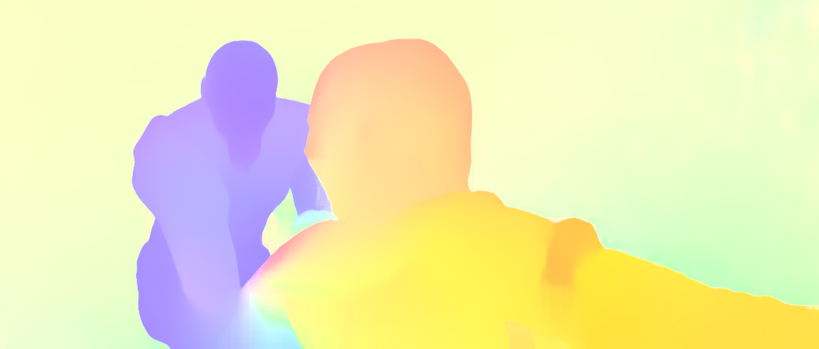} &
  \includegraphics{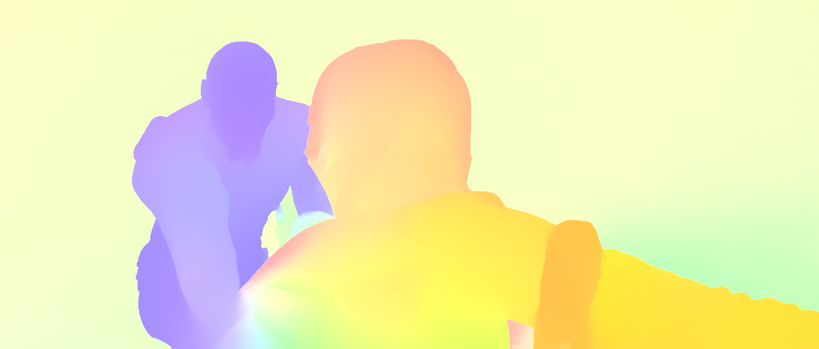}&
  \includegraphics{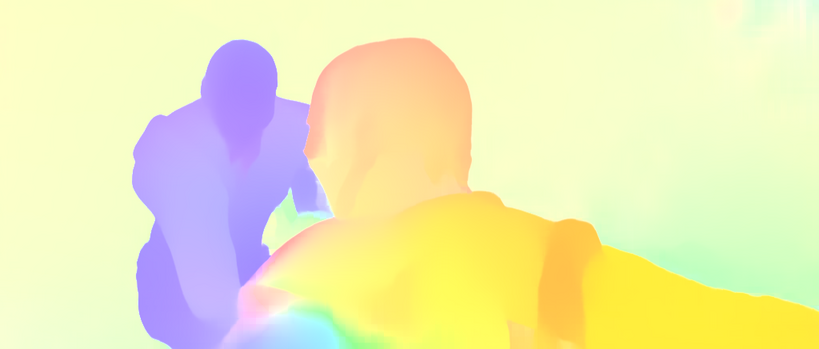} \\
  ``Ambush\_1''& 0 (AEPE $\downarrow$) & 25.69 & 13.06 &   24.12\\
  \\
  \includegraphics{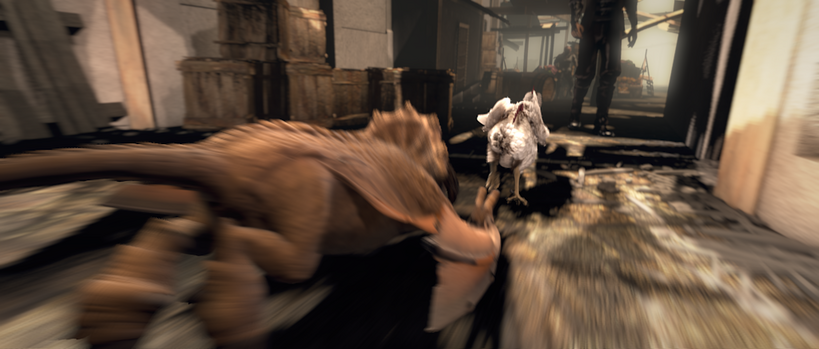}&
  \includegraphics{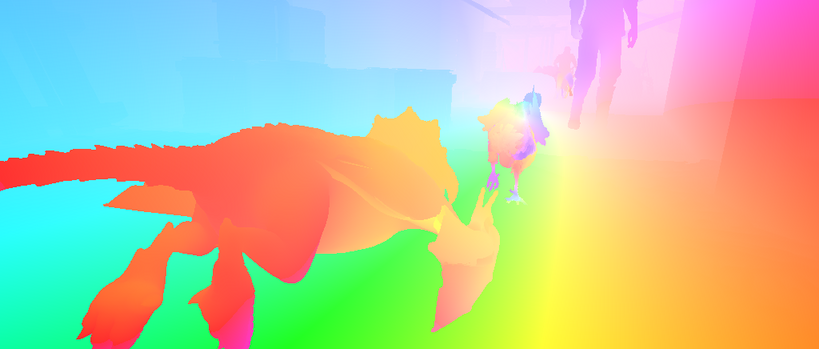}&
  \includegraphics{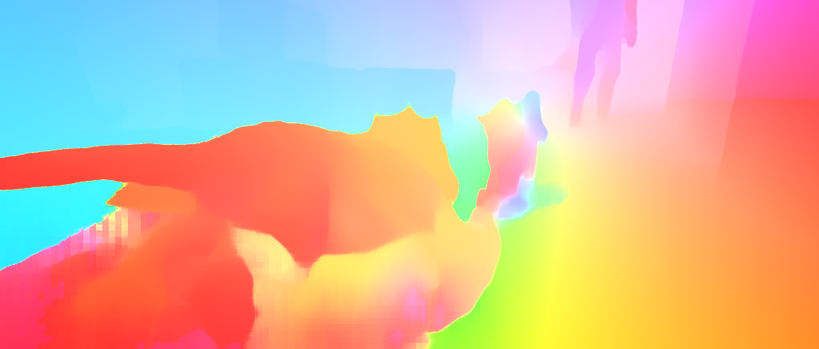} &  
  \includegraphics{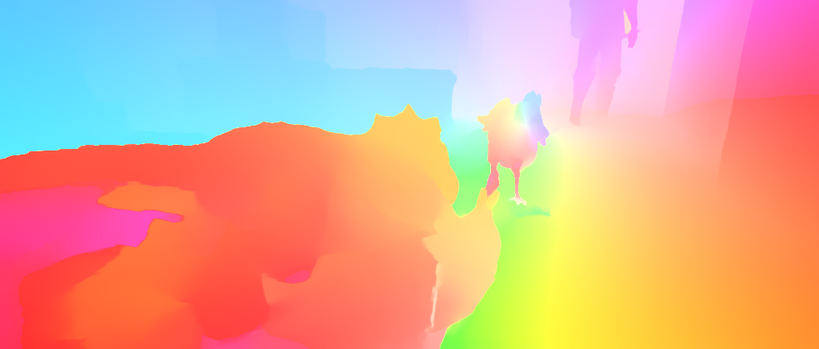}&
  \includegraphics{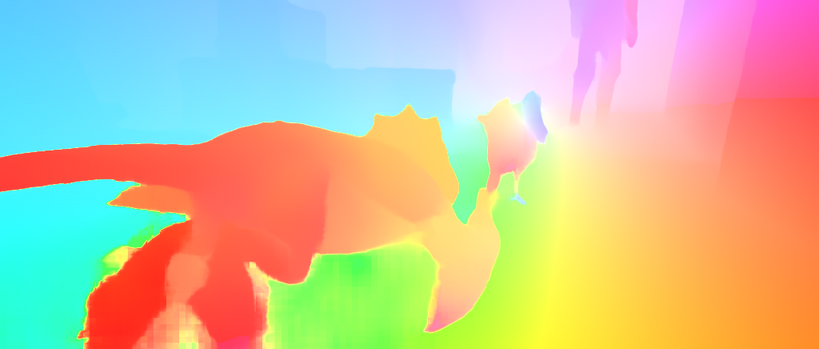}
 \\
 ``Market\_4''& 0 (AEPE $\downarrow$) & 9.07 & 10.38 & 8.18 \\
  \end{tabular}}
  \caption{Visual comparison on two challenging sequences from the Sintel test set. All 2-frame methods make large errors due to heavy snow on ``Ambush\_1'', while RAFT models have larger errors. For the fast moving dragon under motion blur in ``Market\_4'', the newly trained RAFT-it can better resolve the foreground motion from the background than SeparableFlow and the previously trained RAFT~\cite{sun2021autoflow}. 
  }		
\label{fig:sintel:test}
\afterfigure
\end{figure*}

Our newly trained RAFT-it is moderately better than the published RAFT~\cite{sun2021autoflow,teed2020raft}. Among all published {2-frame} methods it is only less accurate than SeparableFlow~\cite{zhang2021separable} while being more than $2\times$ faster in inference. Figure~\ref{fig:sintel:test} visually compares SeparableFlow and our newly trained RAFT on two challenging sequences from Sintel test. RAFT-it makes a larger error on ``Ambush\_1'' under heavy snow, but it correctly predicts the motion of the dragon and the background on ``Market\_4''. To some degree, these comparisons with recent methods compare the effect of innovations on architecture with training techniques, suggesting that there may be large gains for innovations on training techniques. %

\paragraph{KITTI 2015.}
The newly trained PWC-Net-it and IRR-it are substantially better than the respective, published models, with more than 2 percent reduction in average outlier percentage (Fl-all) on the KITTI 2015 benchmark. Both are also more accurate than some more recent models~\cite{mehl2021anisotropic,xu2021flow1d,yang2019volumetric,zhao2020maskflownet}.

\paragraph{Middlebury.}
At the time of writing, our newly trained RAFT-it is ranked first on Middlebury for both end-point and angular errors, with the avg. rank being 1.7 and 3.9, respectively. It is the first deep learning based approach to outperform traditional methods on Middlebury, such as NNF-Local~\cite{chen2013large} (avg. rank 5.8 and 7.4), which had been the top-performing method since 2013. 

\paragraph{VIPER.} Our newly trained RAFT-it obtains 73.6 for the mean weighted area under the
curve (WAUC) over all conditions, \vs 69.5 by  RAFT\_RVC~\cite{tf-raft}.  

\subsection{Higher-resolution Input, Inference Time and Memory}

We perform qualitative evaluations on 2K and 4K resolution inputs from Davis~\cite{Perazzi2016}. For 2K, all models produce similarly high quality flow fields, please see the supplementals for images. In \fig~\ref{fig:pwc:davis-4k}, we  present optical flow results for the newly trained IRR-PWC-it and PWC-Net-it on 4K DAVIS samples. Overall, the flows are comparable, with IRR-PWC-it showing slightly better motion smoothness on the jumping dog (top row in \fig~\ref{fig:pwc:davis-4k}).

Table~\ref{tab:computational_complexity} presents a comparison 
of inference times and memory consumption on an NVIDIA V100 GPU. To account for initial kernel loading, we report the average of 100 runs. For each model, we test three spatial sizes: $1024\mathord\times\mathord448$ (1K), $1920\mathord\times\mathord1080$ (Full HD/2K), and $3840\mathord\times\mathord2160$ (4K). PWC-Net and IRR-PWC show comparable inference time. RAFT, in contrast, is $4.3\mathord\times$ and $14.4\mathord\times$ slower in 1K and 2K, respectively. In terms of memory, PWC-Net and IRR-PWC , again, show comparable performance. The increase in memory usage from 1K to 2K is almost linear for PWC-Net and IRR-PWC. On the other hand, RAFT uses more memory.  Its footprint grows almost quadratically, by $3.8\mathord\times$, from 1K to 2K, and at 4K resolution, RAFT leads to out-of-memory (OOM).

\begin{figure*}[ht!]
    \centering
    \begin{tabular}{ccc}
    First frame & PWC-it & IRR-it \\
    \includegraphics[trim={0 0 0 0},clip,width=0.33\linewidth]{figures/dogs-jump/00033.jpg} & 
    \includegraphics[trim={0 0 0 0},clip,width=0.33\linewidth]{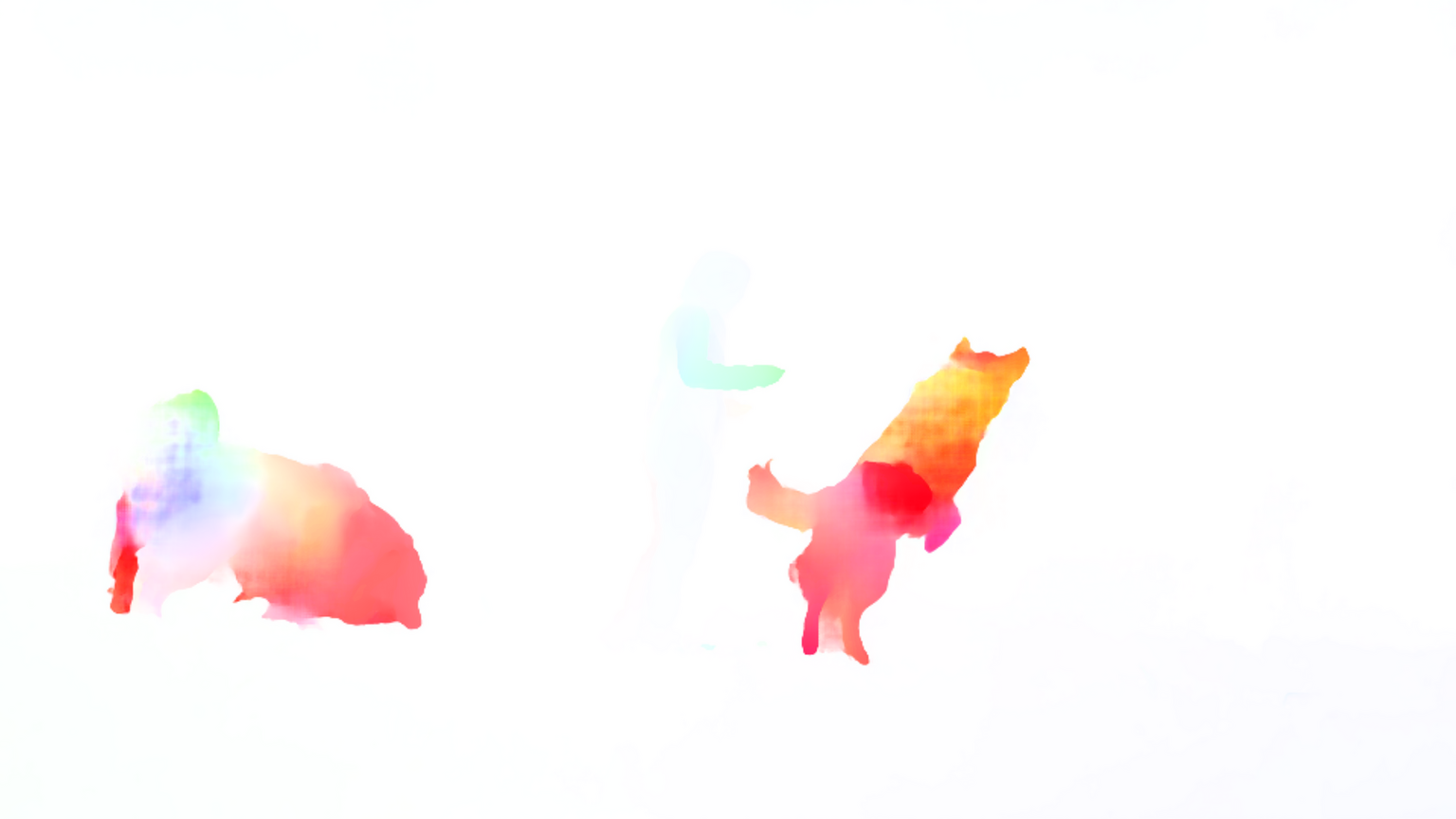} & 
    \includegraphics[trim={0 0 0 0},clip,width=0.33\linewidth]{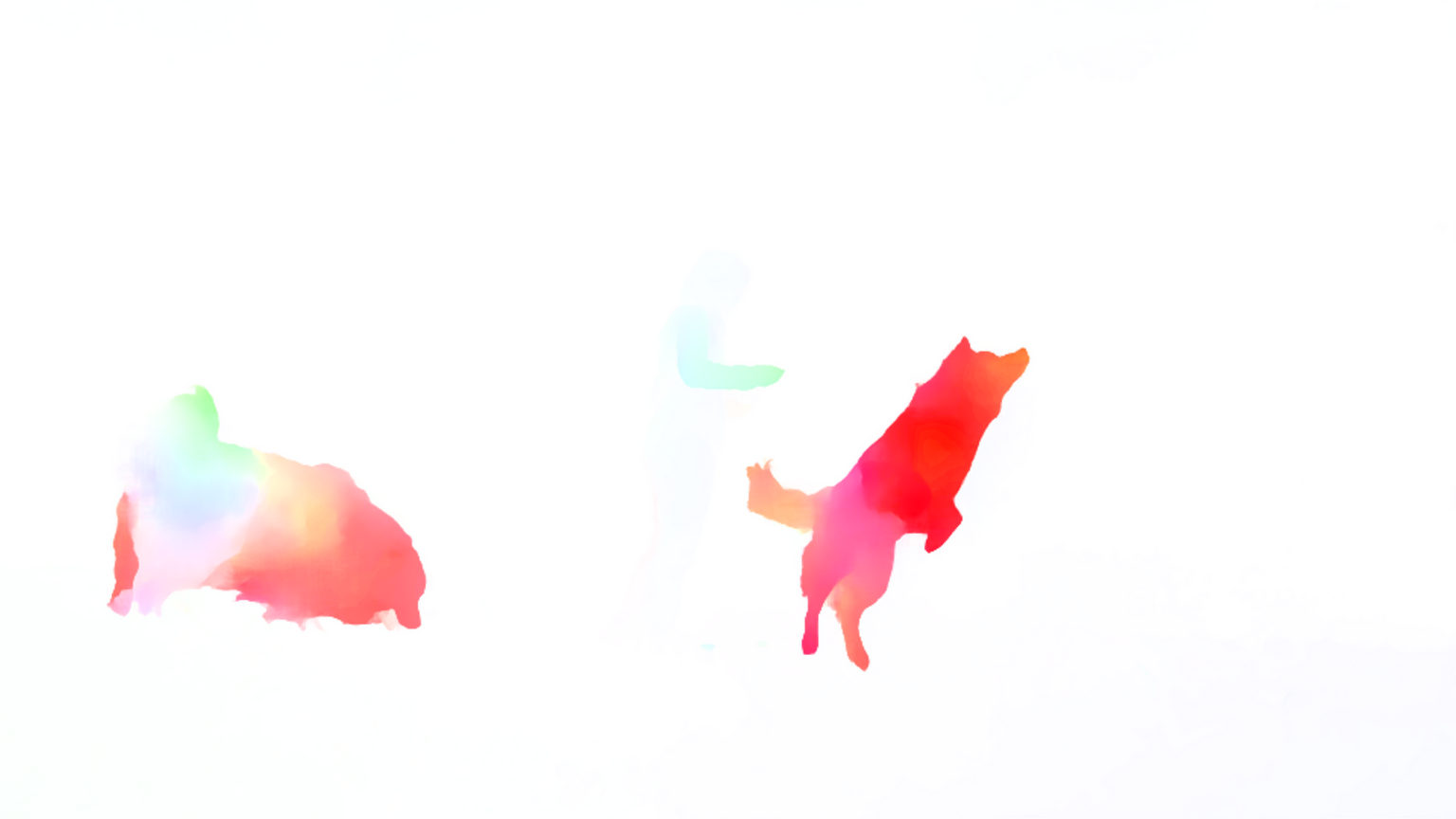} \\ 
    \includegraphics[trim={0 0 0 0},clip,width=0.33\linewidth]{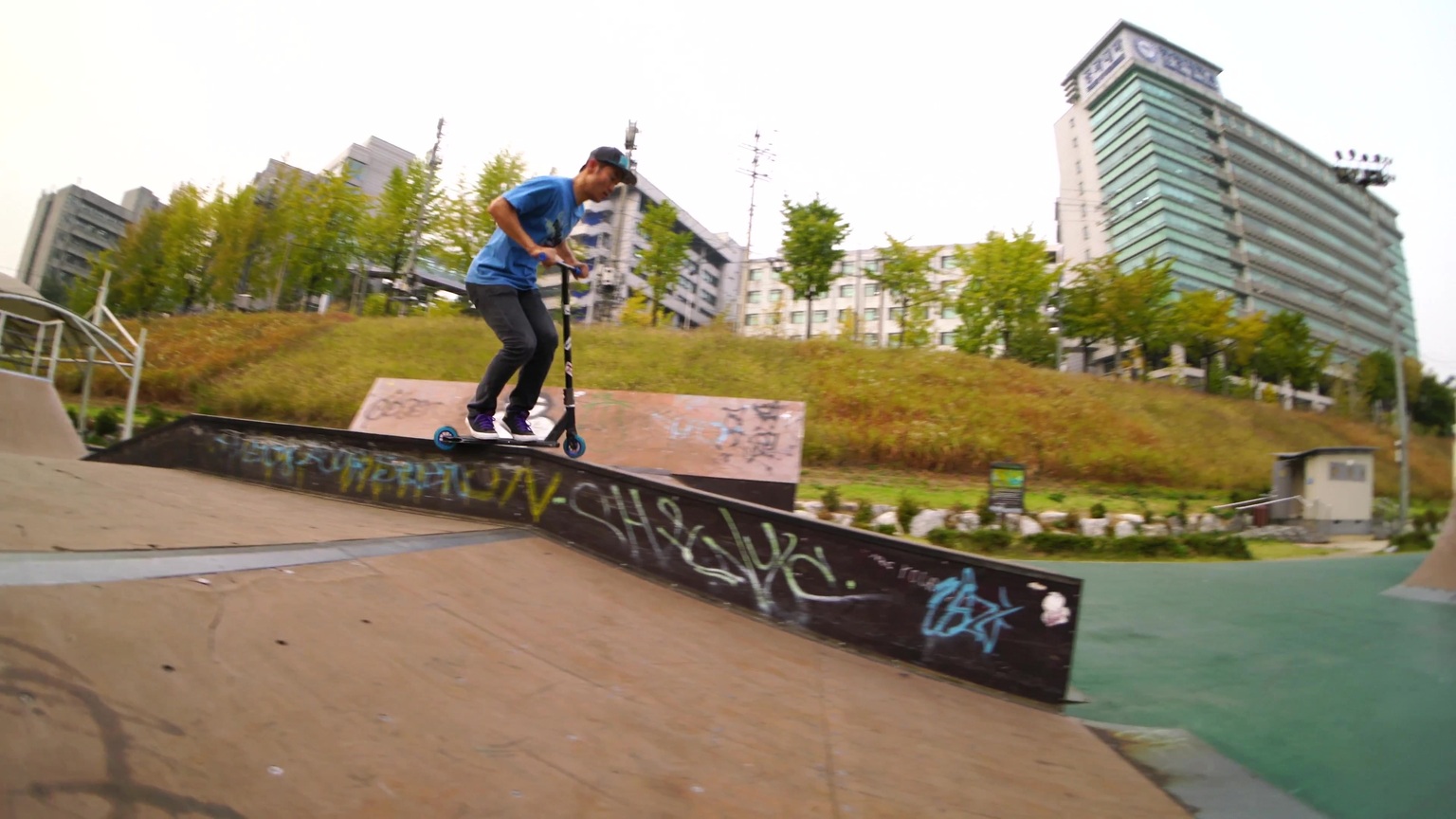} & 
    \includegraphics[trim={0 0 0 0},clip,width=0.33\linewidth]{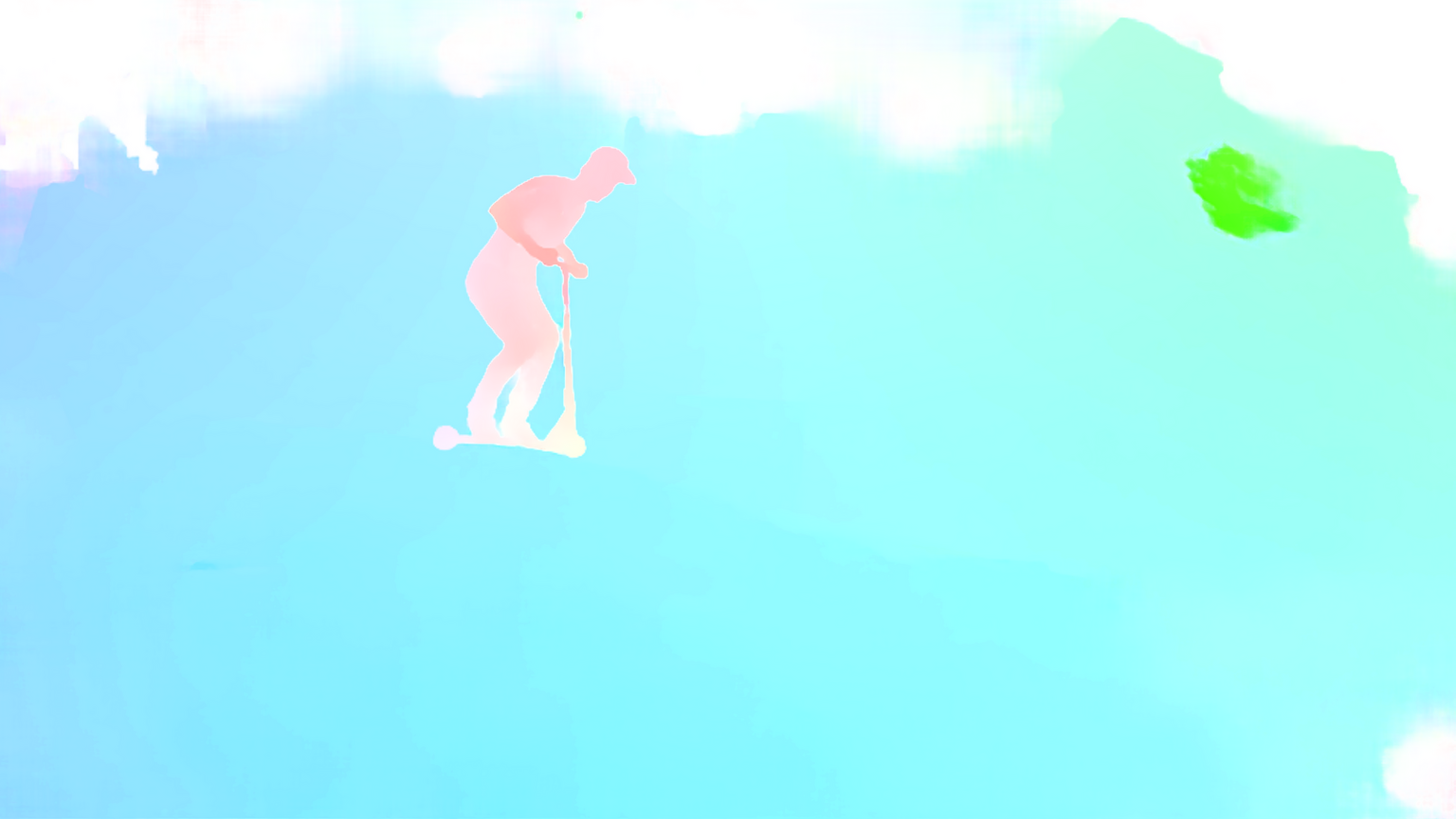} & 
    \includegraphics[trim={0 0 0 0},clip,width=0.33\linewidth]{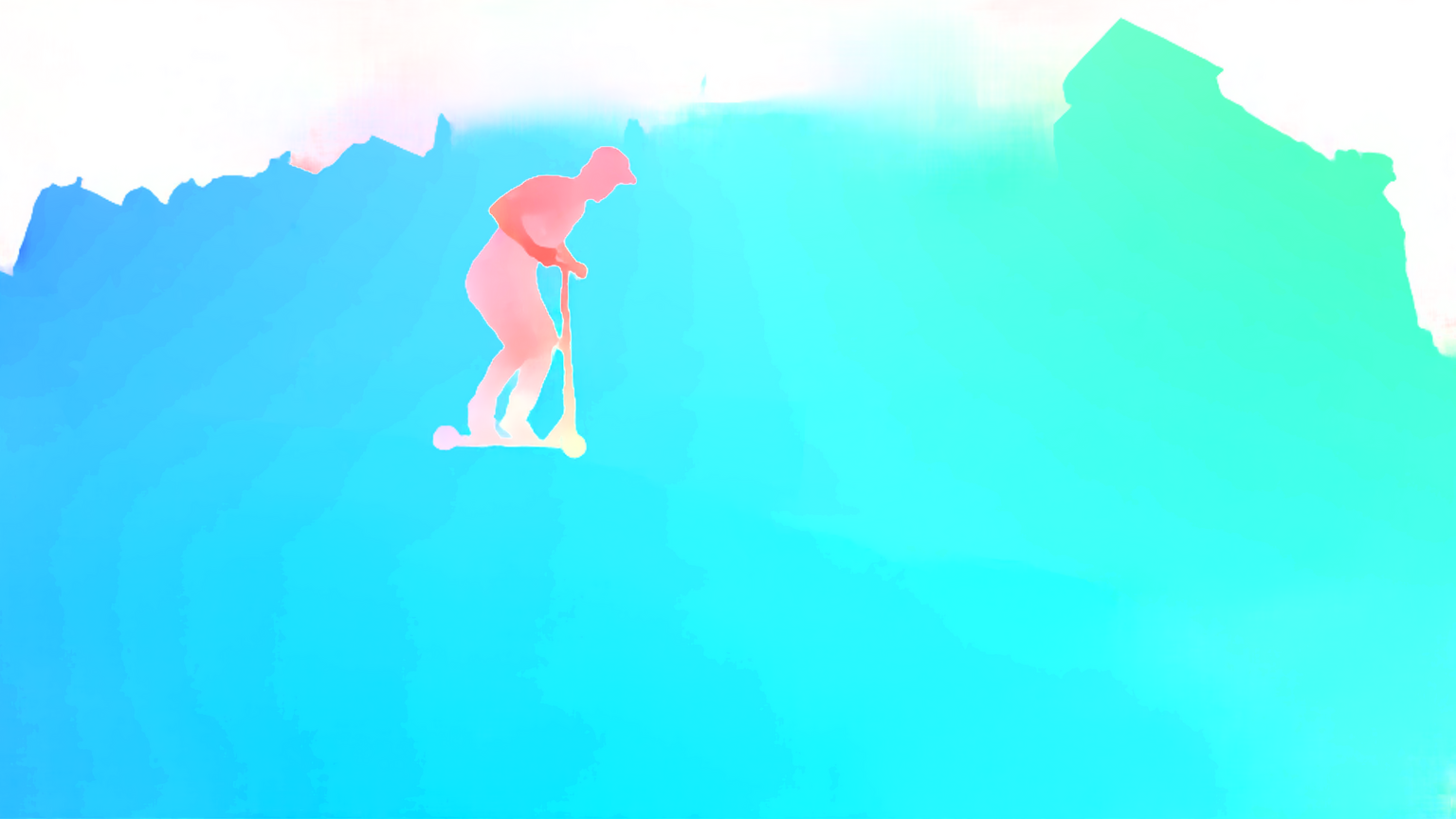} \\ 
    \end{tabular}
	\caption{Visual results on \textbf{Davis 4K}. We show only PWC-Net-it and IRR-PWC-it results 
	since RAFT runs out of memory on the 16GB GPU.}
    \label{fig:pwc:davis-4k}
    \afterfigure
\end{figure*}

\begin{table}[htbp!]
\small
\centering
\begin{tabular}{l c c c | c c c}
    & \multicolumn{3}{c}{Inference Time (msec)$\downarrow$} & \multicolumn{3}{c}{Peak Memory (GB)$\downarrow$} \\
    \hline
    & \small{$1024\mathord\times\mathord448$} & Full HD & 4K & $1024\times448$ & Full HD & 4K \\
    \hline
     PWC-Net & 20.61 & 28.77 & 63.31 & 1.478 & 2.886 & 7.610 \\
     IRR-PWC & 24.71 & 33.67 & 57.59 & 1.435 & 2.902 & 8.578 \\
     RAFT & 107.38 & 499.63 &  n/a & 2.551 & 9.673 & OOM \\
    \bottomrule
\end{tabular}
\caption{Inference time and memory usage for $1024\mathord\times448$, Full HD ($1920\mathord\times1080$) and 4K ($3840\mathord\times2160$) frame sizes, averaged over 100 runs on an NVIDIA V100 GPU. %
}
\label{tab:computational_complexity}
\aftertable
\end{table}

\subsection{Discussion}

\paragraph{What makes RAFT better than PWC-Net?}
{
Our results show that several factors contribute to the performance gap between the published RAFT (5.10\% Fl-all on KITTI 2015, see Table~\ref{tab:benchmark}) and PWC-Net (7.72\%) methods, including training techniques, datasets and architecture innovations. 
Recent training techniques and datasets significantly improve PWC-Net (5.54\%) and IRR-PWC (5.73\%). The newly trained models are competitive with published RAFT (5.10\%) performance while maintaining their advantages in speed and memory requirements during inference.  
These insights also yield a newly trained RAFT-it model that sets a new state of the art on Middlebury at the time of writing.
We conclude that innovations on training techniques and datasets 
are another fruitful path to performance gains, for both old and new optical flow architectures.
After compensating for the differences in training techniques and datasets, we can identify the true performance gap between PWC-Net and RAFT that is solely due to architecture innovations ($5.54$\% vs. $4.31\%$ Fl-all on KITTI 2015).
Future work should examine which specific architecture elements of RAFT are critical, and whether they may be transferable to other models. 
}

\paragraph{No model to rule all.} 
{Our study also shows that there are several factors to consider when choosing an optical flow model, 
including flow accuracy, training time, inference time,  memory cost and application scenarios. RAFT has the highest accuracy and faster convergence in training, but is slower at test time and has a high memory footprint. %
PWC-Net and IRR-PWC are more appealing for applications that require fast inference, low memory cost and high-resolution input. 
PWC-Net may be suitable for applications with small motions.
Every model entails trade-offs between different requirements; no 
single model is superior on all metrics.
Thus, researchers may wish to focus on specific metrics for improvement,
thereby providing practitioners with more options. 
}

\section{Conclusions}

We have evaluated three prominent optical flow architectures with improved training protocols and observed surprising and significant performance gains. 
The newly trained PWC-Net-it and IRR-PWC-it are more accurate than the more recent Flow1D model on KITTI 2015, while being about $3\times$ faster during inference.
Our newly trained RAFT-it sets a new state of the art and is the first deep learning approach to outperform traditional methods on the Middlebury benchmark.
These results demonstrate the benefits of decoupling the contributions of model architectures, training techniques, and datasets to understand the sources of performance gains.

\bibliographystyle{splncs04}
\bibliography{main}

\appendix
\onecolumn

The main paper includes only several examples on the Davis datasets due to space limits. Here we provide more visual examples to more comprehensively evaluate these models visually. We also include screenshots that indicate how our method does on public benchmarks and  detailed results on these benchmarks. Throughout the document, we add  ``-it'' to each method to denote our newly trained model, where ``it'' stands for improved training.

\section{Screenshots of Public Benchmarks }

\begin{figure*}[!htb]
    \centering
    \small
    \includegraphics[width=\linewidth]{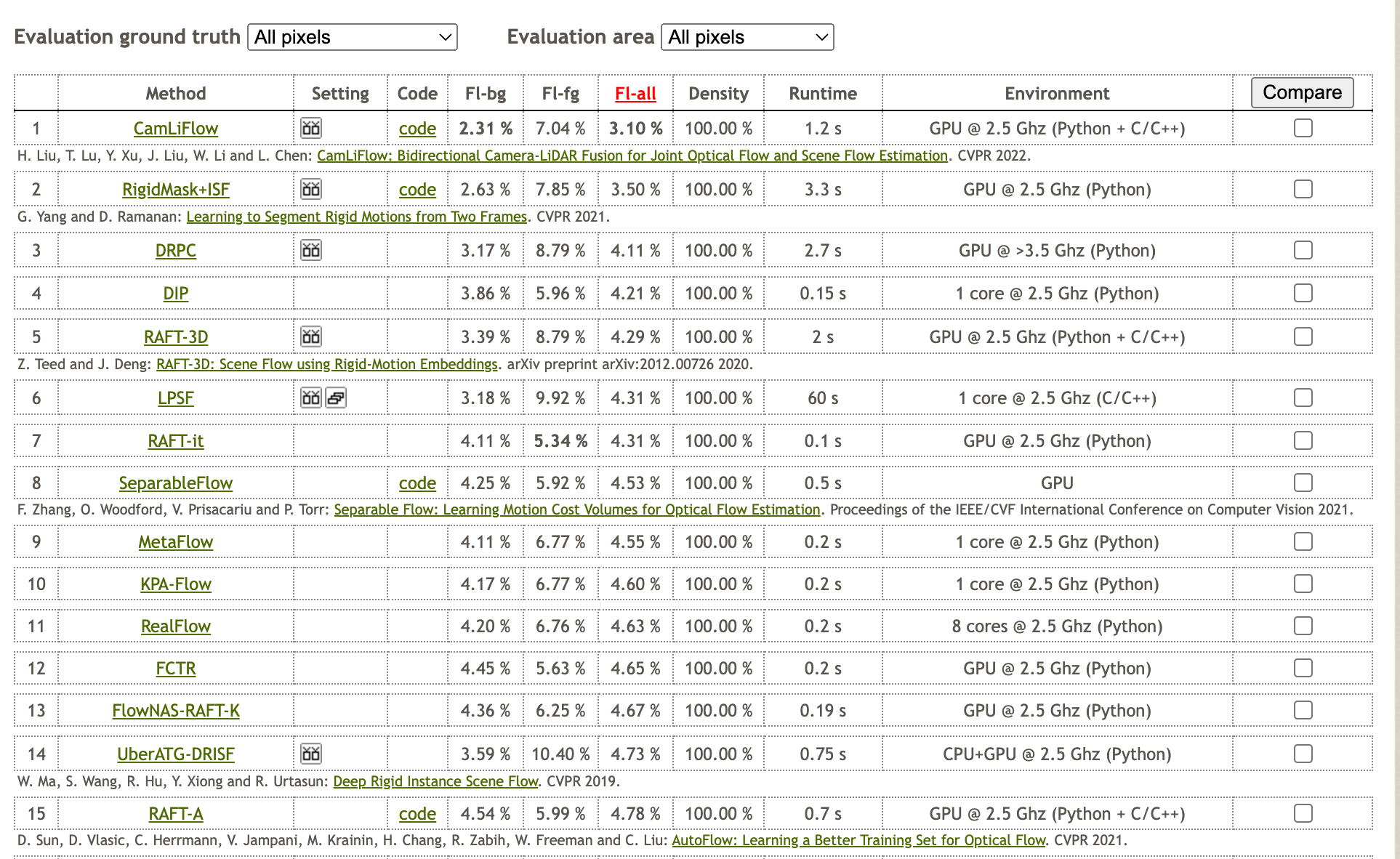}
	\caption{Screenshot of KITTI 2015 public benchmark. We name our newly trained RAFT as RAFT-it,  ``it'' stands for improved training. }
    \label{fig:kitti:screen}
    \vspace{-2ex}
\end{figure*}

\begin{figure*}[!htb]
    \centering
    \small
    \includegraphics[width=\linewidth]{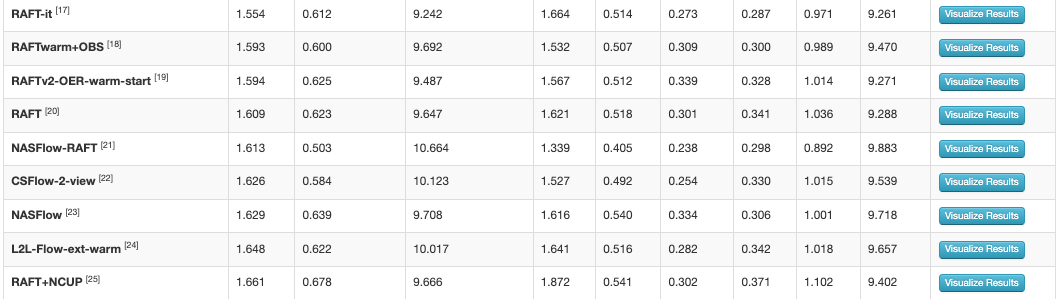}
	\caption{Screenshot of Sintel clean public benchmark. We name our newly trained RAFT as RAFT-it,  ``it'' stands for improved training. }
    \label{fig:sintel:clean:screen}
    \vspace{-2ex}
\end{figure*}

\begin{figure*}[!htb]
    \centering
    \small
        \includegraphics[width=\linewidth]{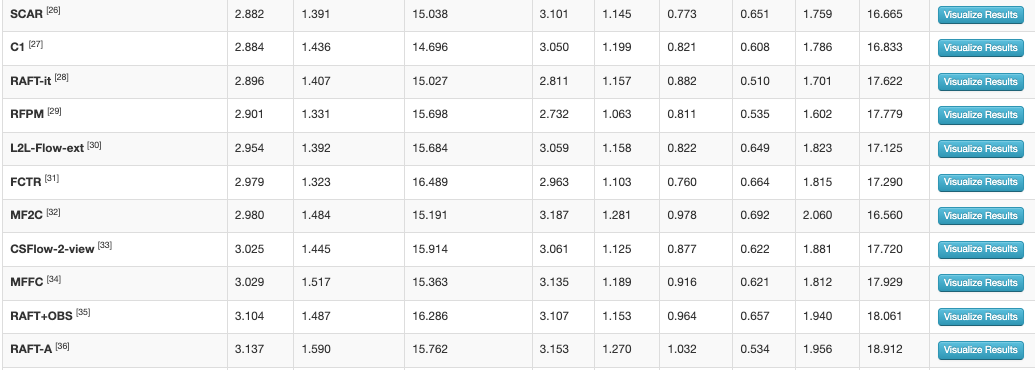}
	\caption{Screenshot of Sintel final public benchmark. We name our newly trained RAFT as RAFT-it,  ``it'' stands for improved training. RAFT-it is only slightly worse than SeparableFlow on Sintel.final among all published two-frame methods.}
    \label{fig:sintel:final:screen}
    \vspace{-2ex}
\end{figure*}

\begin{figure*}[!htb]
    \centering
    \small
        \includegraphics[width=\linewidth]{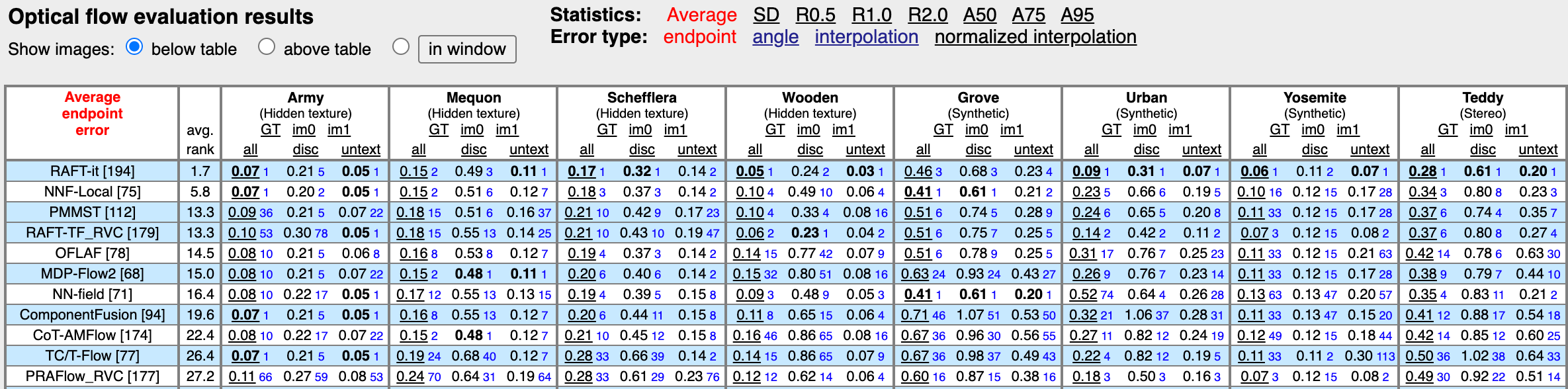}
	\caption{Screenshot of Middlebury public benchmark (AEPE). We name our newly trained RAFT as RAFT-it,  ``it'' stands for improved training. RAFT-it sets a new state of the art on Middlebury.}
    \label{fig:middlebury:aepe:screen}
    \vspace{-2ex}
\end{figure*}

Table~\ref{tab:ft:sintel:details} summarizes the detailed results by previously published and our newly trained models. 
The newly trained models are more accurate than previously models regardless of occlusions (unmatched), distance to motion boundaries, and speed.

\begin{table}[h]
\begin{center}
\small
\begin{tabular}{ lccccccccc } 
{\bf Model} & {\bf all}	& {\bf match} & {\bf unmatch}	& {\bf d0-10} &	{\bf d10-60} &	{\bf d60-140} &	{\bf s0-10} & {\bf s10-40} &	{\bf s40+} \\ \hline\hline
PWC-Net & 4.60	&2.25	&23.70	&4.78	&2.05	&1.23	&0.95	&2.98	&26.62 \\
PWC-Net-it (ours) & 3.68 & 1.82 & 18.87 & 3.47 & 1.39 & 1.18 & 0.62 & 1.96 & 23.07  \\ \hline
IRR-PWC & 4.58	&2.15	&24.36	&4.17	&1.84	&1.29	&0.71	&2.42	&29.00	\\
IRR-PWC-it (ours) & 3.56 & 1.83 & 17.54 & 3.67 & 1.40 & 1.16 & 0.63 & 2.04 & 21.63 \\ \hline
RAFT~\cite{sun2021autoflow} &3.14	&1.59	&15.76	&3.15	&1.27	&1.03	&0.53	&1.96	&18.91\\
RAFT-it (ours) &2.90	&1.41	&15.03	&2.81	&1.16	&0.88	&0.51	&1.70	&17.62 \\
\end{tabular}
\caption{Detailed analysis of AEPE on Sintel test set. ``it'' stands for improved training. 
}
\label{tab:ft:sintel:details}
\end{center}
\end{table}

Table~\ref{tab:kitti:details} summarizes the detailed results on KITTI for the previously best published and the newly trained models. The newly trained models are generally better than the previously trained models. The only exception is the foreground regions for IRR-PWC. Note that the original IRR-PWC implementation computes bidirectional flow, reasons about occlusions, and uses a bilateral refinement, which may help the foreground objects. Our newly trained IRR-PWC is a straightforward modification of PWC-Net and is more lightweight without these sophisticated modules. 

\begin{table}[ht]
\begin{center}
\begin{tabular}{l|c c  c c c c c}
\multirow{2}{*}{{\bf Model}} & \multicolumn{3}{c}{All} & \multicolumn{3}{c}{Occ} \\
 & {\bf Fl-bg} & {\bf Fl-fg} & {\bf Fl-all} & {\bf Fl-bg} & {\bf Fl-fg} & {\bf Fl-all} \\ \hline
PWC-Net~\cite{sun2018pwc}  & 9.66 \% & 9.31 \% & 9.60 \% & 6.14 \% & 5.98 \% & 6.12 \%\ \\
PWC-Net~\cite{sun2019models} & 7.69 \% & 7.88 \% & 7.72 \% & 4.91 \% & 4.88 \% & 4.91\%  \\
PWC-Net-it (ours) & 5.18 \% & 7.36 \% & 5.54 \% & 3.41 \% & 4.90 \% & 3.68 \%\\ \hline
IRR-PWC & 7.68 \% & 7.52 \% & 7.65 \% & 4.92 \% & 4.62 \% & 4.86 \%\\
IRR-PWC-it (ours) & 5.12 \% & 8.82 \% & 5.73 \%  & 3.47 \% & 5.95 \% & 3.92 \%\\  \hline
RAFT~\cite{teed2020raft} & 4.74 \% & 6.87 \% & 5.10 \% & 2.87 \% & 3.98 \% & 3.07 \%\\
RAFT~\cite{sun2021autoflow} & 4.54 \% & 5.99 \% & 4.78  \% & 3.01 \% & 3.17 \% & 3.04\%\\
RAFT-it (ours) &  4.11 \% & 5.34 \% & 4.31 \% & 2.68 \% & 2.77 \% & 2.70 \%\\ 
\end{tabular}
\caption{Detailed performance on KITTI 2015 test set. ``it'' stands for improved training. }
\label{tab:kitti:details}
\end{center}
\end{table}

\section{More Visual Comparisons }

\subsection{PWC-it, IRR-it, and RAFT-it on Davis 2K}

\begin{figure}[!htb]
\begin{tabular}{cc}
\includegraphics[width=.5\linewidth]{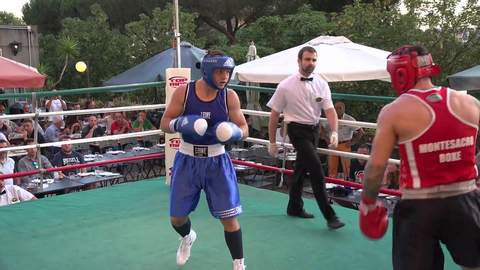}
& \includegraphics[width=.5\linewidth]{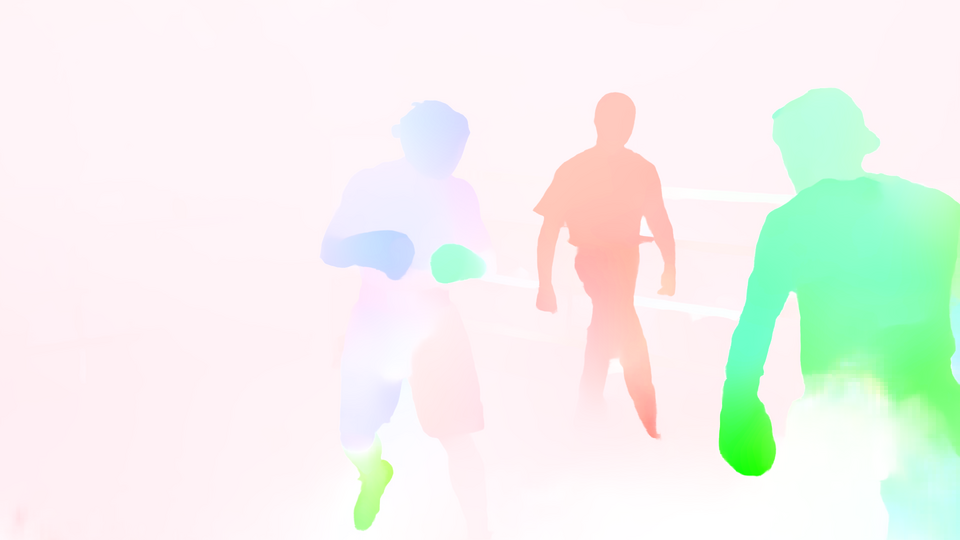} \\
{\scriptsize Frame 0} & {\scriptsize RAFT-it} \\ 
\includegraphics[width=.5\linewidth]{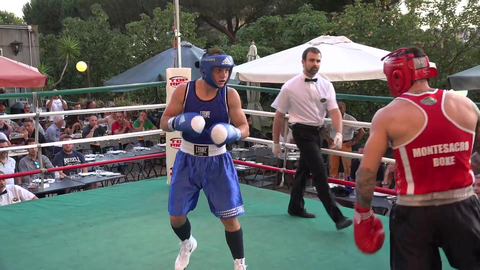}
& {\includegraphics[width=.5\linewidth]{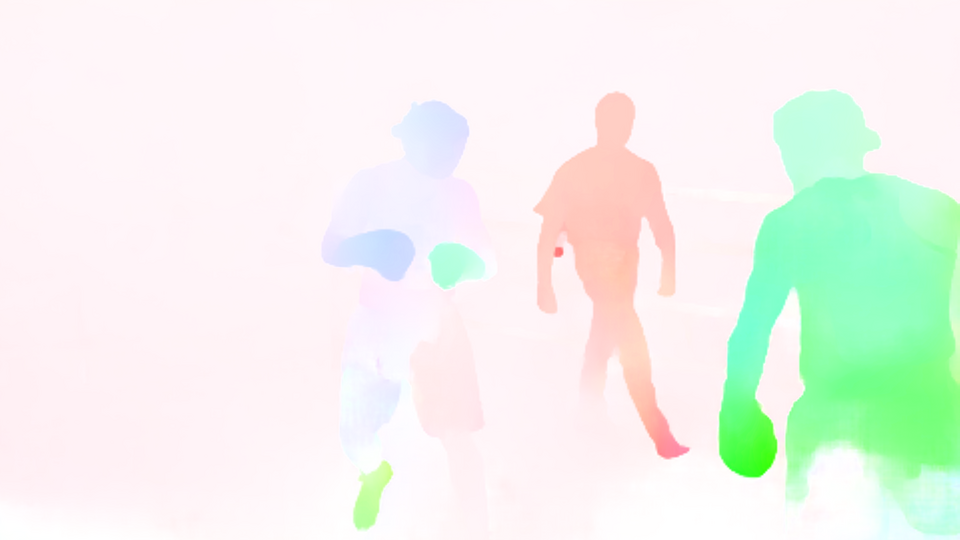}} \\
{\scriptsize Frame 1} & {\scriptsize IRR-it} \\ 
& {\includegraphics[width=.5\linewidth]{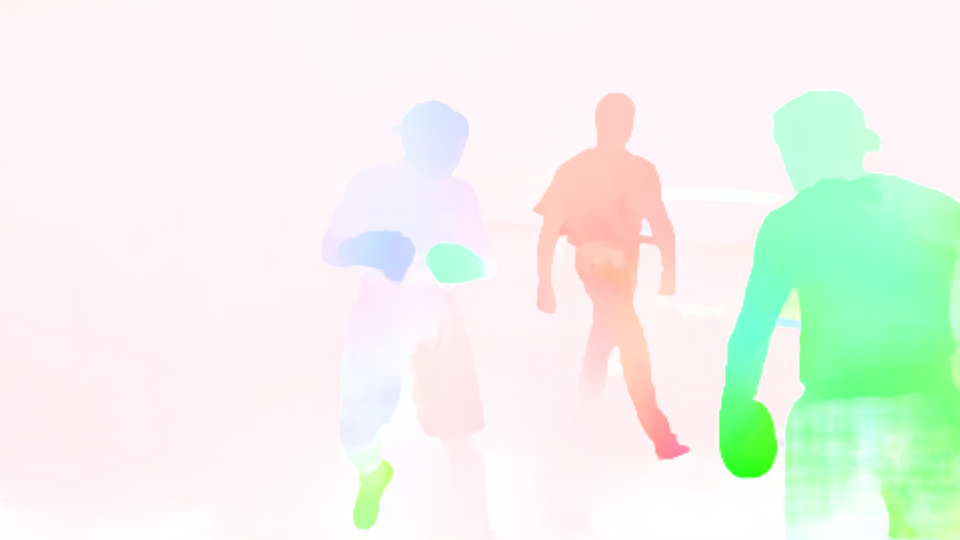}} \\
 & {\scriptsize PWC-Net-it} \\ 
\end{tabular}
\caption{PWC-it, IRR-it, and RAFT-it on Davis 2K images.  }
\label{fig:2k_1}
\end{figure}

\begin{figure}[!htb]
\begin{tabular}{cc}
\includegraphics[width=.5\linewidth]{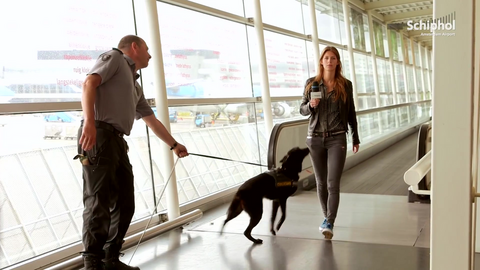}
& \includegraphics[width=.5\linewidth]{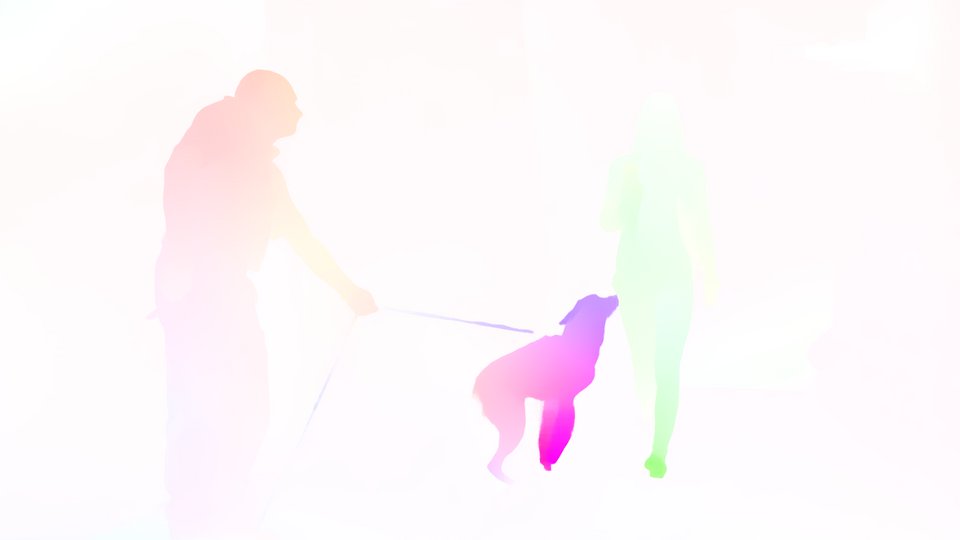} \\
{\scriptsize Frame 0} & {\scriptsize RAFT-it} \\ 
\includegraphics[width=.5\linewidth]{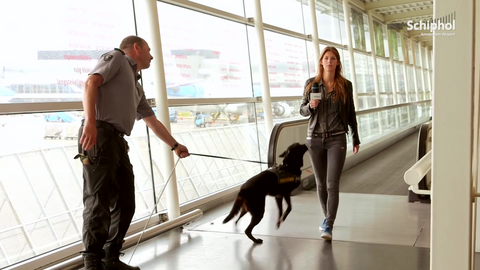}
& {\includegraphics[width=.5\linewidth]{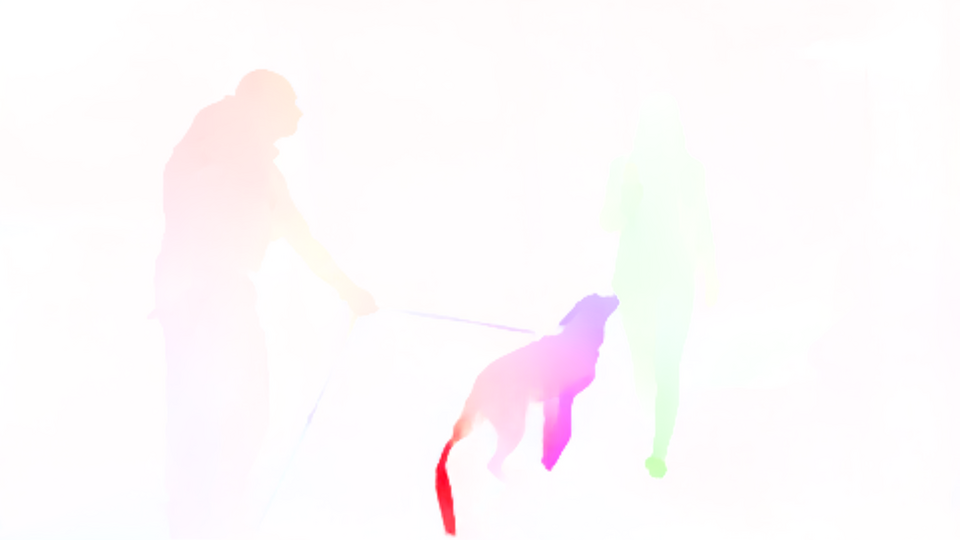}} \\
{\scriptsize Frame 1} & {\scriptsize IRR-it} \\ 
& {\includegraphics[width=.5\linewidth]{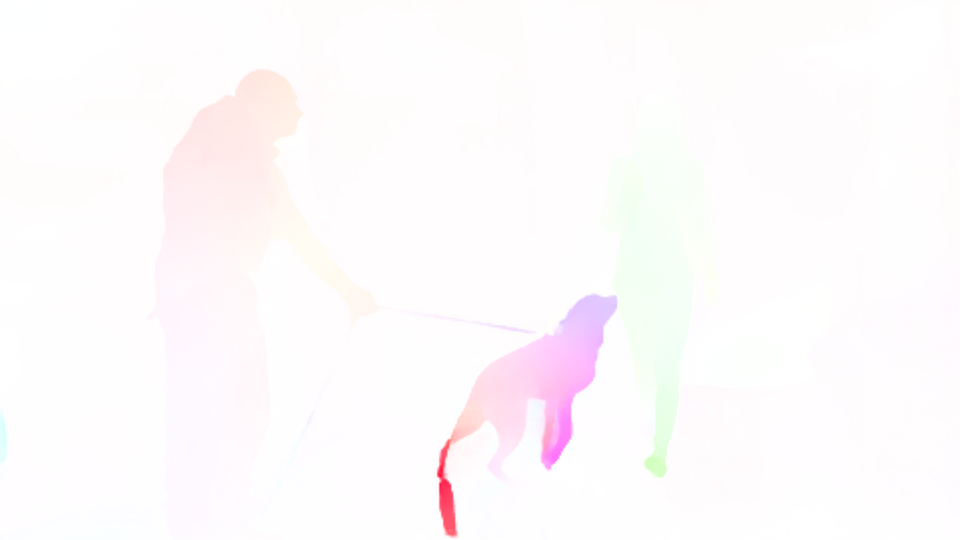}} \\
 & {\scriptsize PWC-Net-it} \\ 
\end{tabular}
\caption{PWC-it, IRR-it, and RAFT-it on Davis 2K images.  }
\label{fig:2k_3}
\end{figure}

\begin{figure}[!htb]
\begin{tabular}{cc}
\includegraphics[width=.5\linewidth]{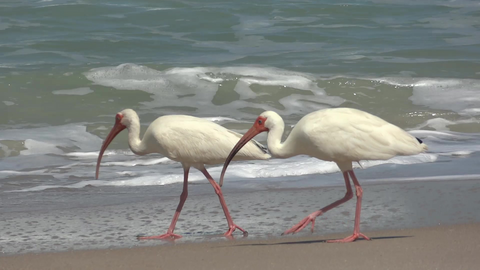}
& \includegraphics[width=.5\linewidth]{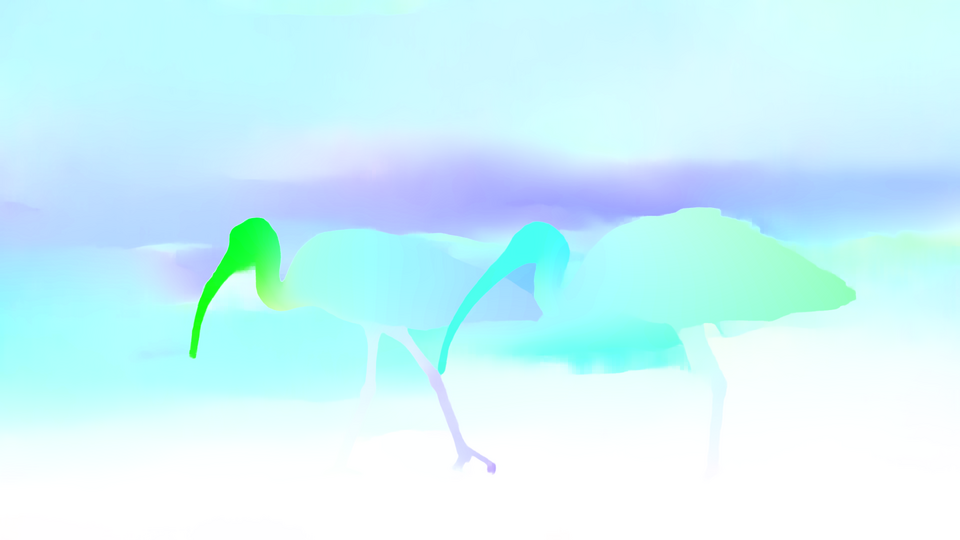} \\
{\scriptsize Frame 0} & {\scriptsize RAFT-it} \\ 
\includegraphics[width=.5\linewidth]{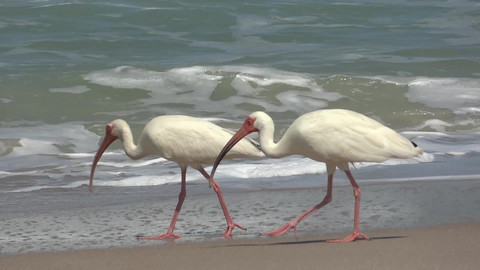}
& {\includegraphics[width=.5\linewidth]{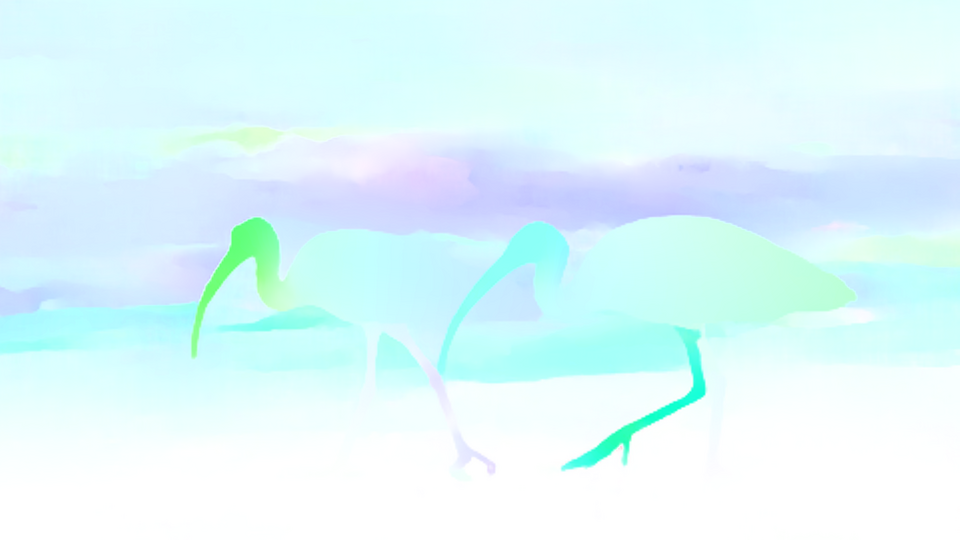}} \\
{\scriptsize Frame 1} & {\scriptsize IRR-it} \\ 
& {\includegraphics[width=.5\linewidth]{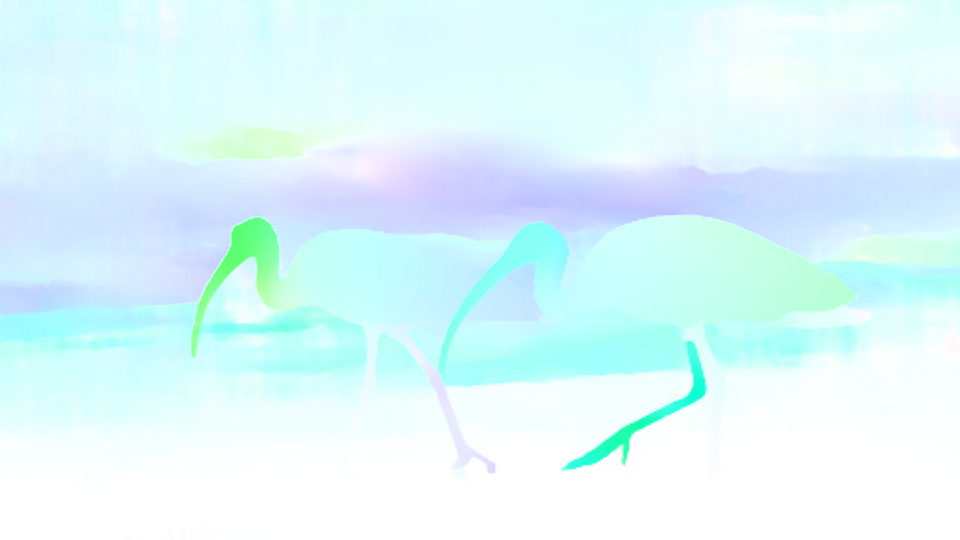}} \\
 & {\scriptsize PWC-Net-it} \\ 
\end{tabular}
\caption{PWC-it, IRR-it, and RAFT-it on Davis 2K images.  }
\label{fig:2k_4}
\end{figure}

\begin{figure}[!htb]
\begin{tabular}{cc}
\includegraphics[width=.5\linewidth]{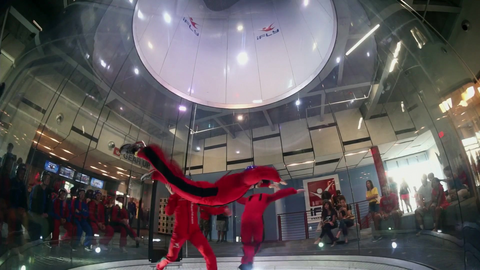}
& \includegraphics[width=.5\linewidth]{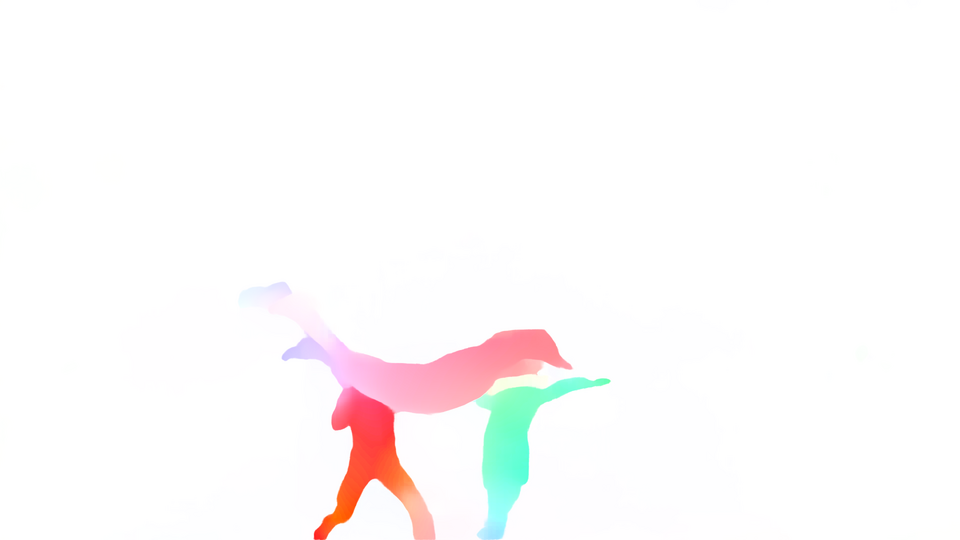} \\
{\scriptsize Frame 0} & {\scriptsize RAFT-it} \\ 
\includegraphics[width=.5\linewidth]{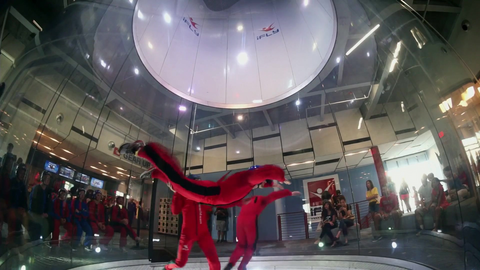}
& {\includegraphics[width=.5\linewidth]{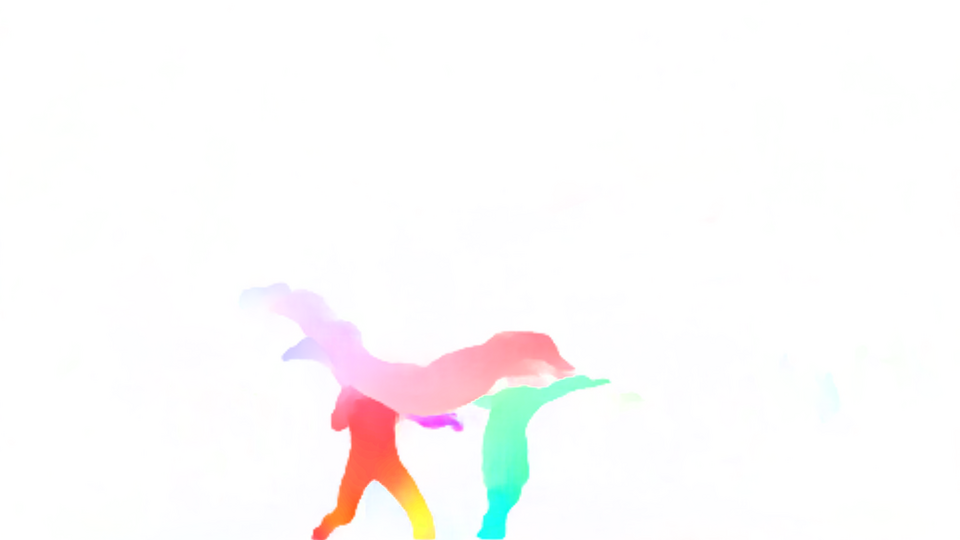}} \\
{\scriptsize Frame 1} & {\scriptsize IRR-it} \\ 
& {\includegraphics[width=.5\linewidth]{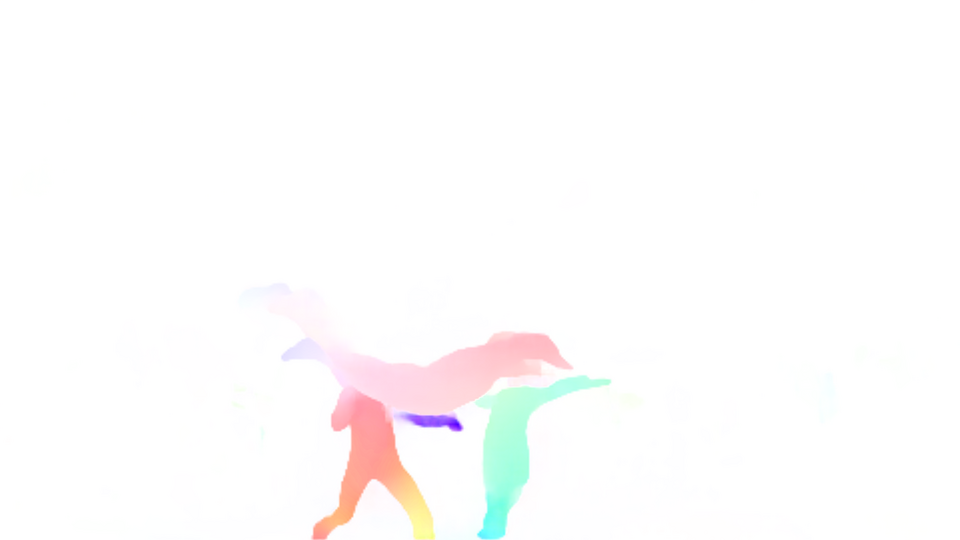}} \\
 & {\scriptsize PWC-Net-it} \\ 
\end{tabular}
\caption{PWC-it, IRR-it, and RAFT-it on Davis 2K images.  }
\label{fig:2k_5}
\end{figure}

In this subsection, we include 4 examples of our improved training models on Davis 2K images: Figures \ref{fig:2k_1}, \ref{fig:2k_3}, \ref{fig:2k_4}, and \ref{fig:2k_5}. 

\subsection{PWC-it, IRR-it, and RAFT-it (down-up) on Davis 4K}

\begin{figure}[!htb]
\begin{tabular}{cc}
\includegraphics[width=.5\linewidth]{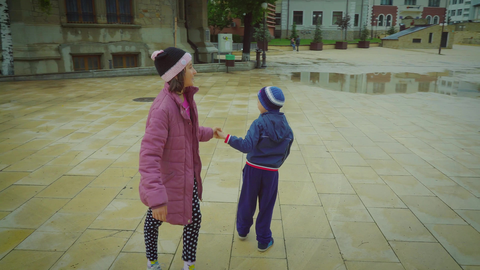}
& \includegraphics[width=.5\linewidth]{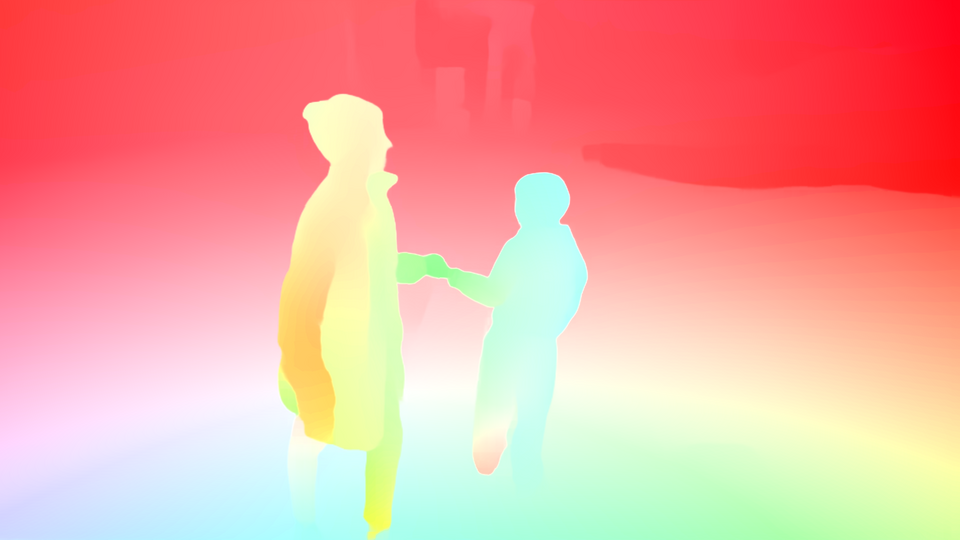} \\
{\scriptsize Frame 0} & {\scriptsize RAFT-it (down-up)} \\ 
\includegraphics[width=.5\linewidth]{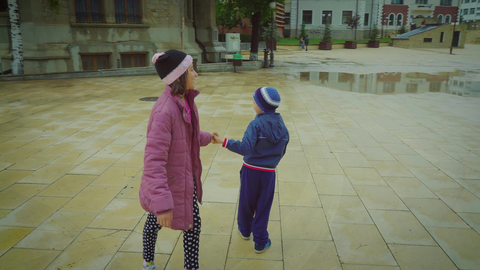}
& {\includegraphics[width=.5\linewidth]{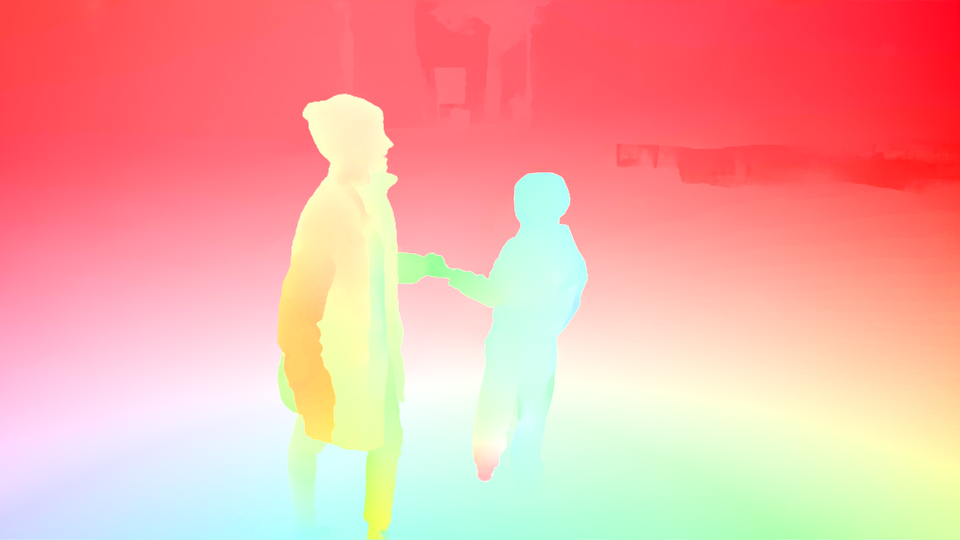}} \\
{\scriptsize Frame 1} & {\scriptsize IRR-it} \\ 
& {\includegraphics[width=.5\linewidth]{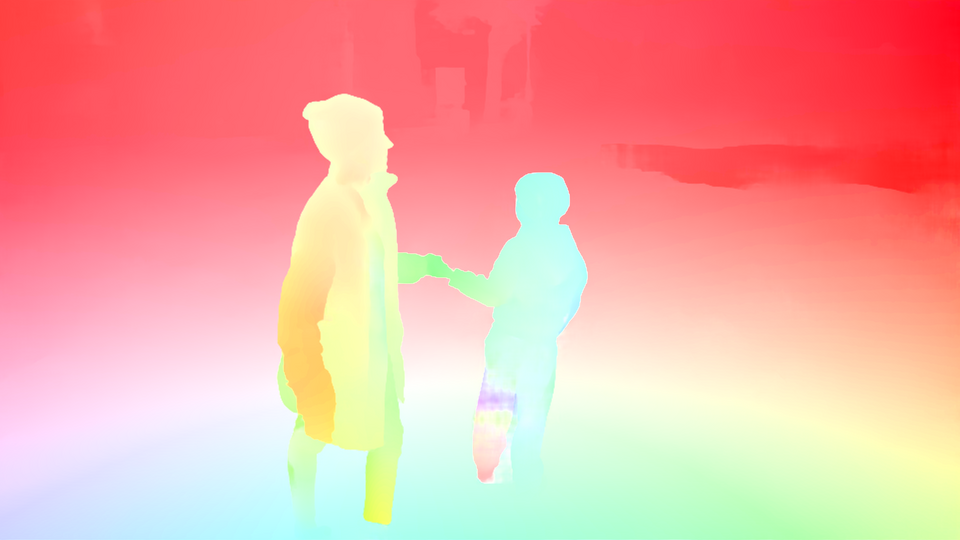}} \\
 & {\scriptsize PWC-Net-it} \\ 
\end{tabular}
\caption{PWC-it, IRR-it, and RAFT-it (down-up) on Davis 4K images. Note that RAFT-it requires the input images to be downsampled due to memory requirements and then the flow upsampled. Note the higher level of detail in IRR-it and PWC-Net-it: the clothes on the person on the left. }
\label{fig:4k_5}
\end{figure}

\begin{figure}[!htb]
\begin{tabular}{cc}
\includegraphics[width=.5\linewidth]{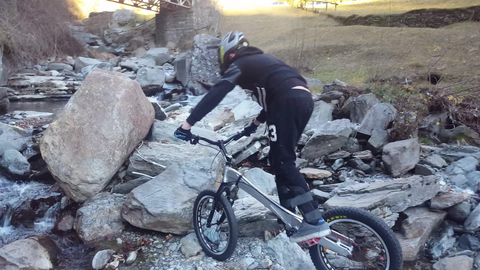}
& \includegraphics[width=.5\linewidth]{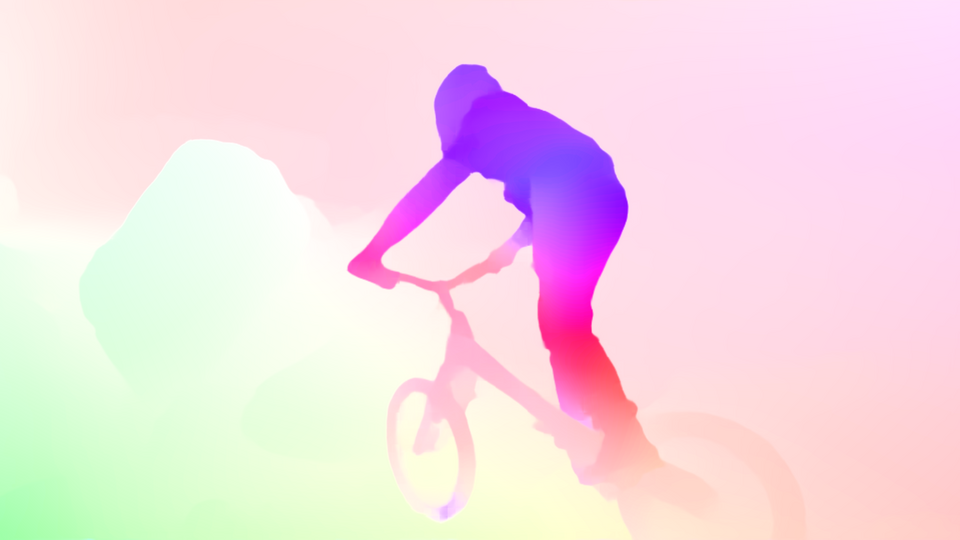} \\
{\scriptsize Frame 0} & {\scriptsize RAFT-it (down-up)} \\ 
\includegraphics[width=.5\linewidth]{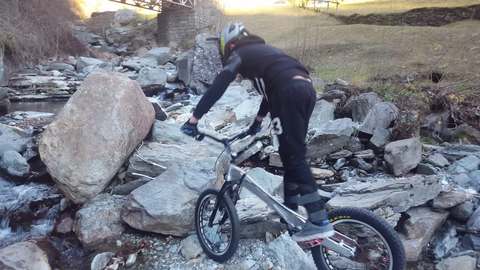}
& {\includegraphics[width=.5\linewidth]{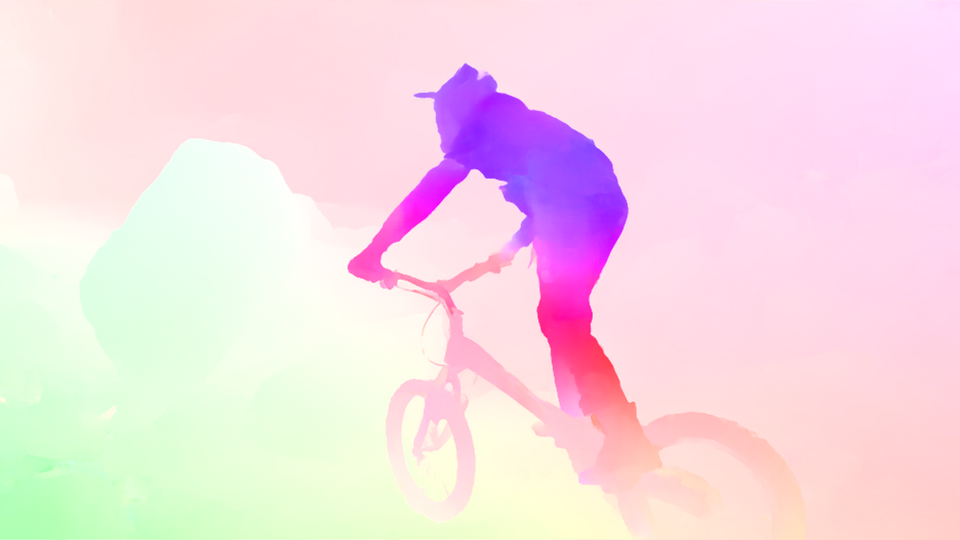}} \\
{\scriptsize Frame 1} & {\scriptsize IRR-it} \\ 
& {\includegraphics[width=.5\linewidth]{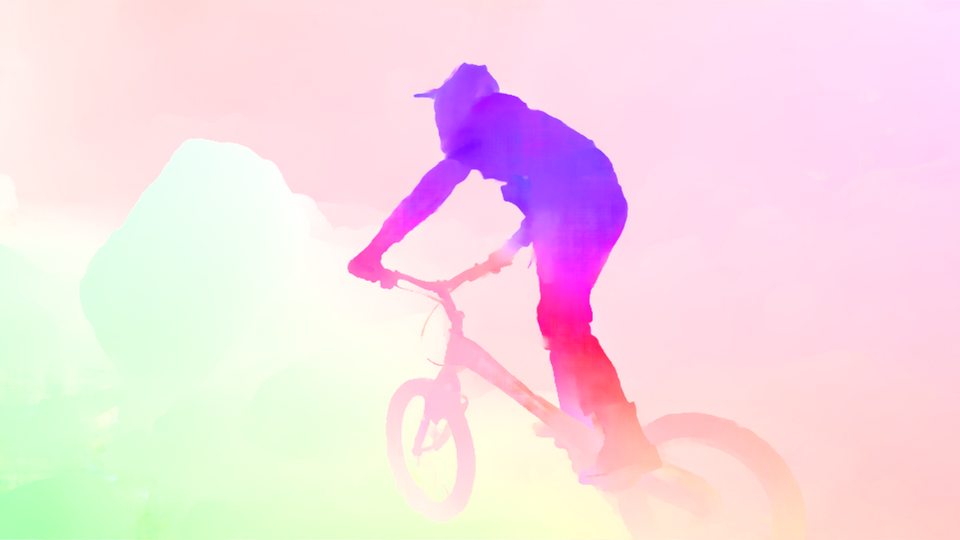}} \\
 & {\scriptsize PWC-Net-it} \\ 
\end{tabular}
\caption{PWC-it, IRR-it, and RAFT-it (down-up) on Davis 4K images. Note that RAFT-it requires the input images to be downsampled due to memory requirements and then the flow upsampled. Note the higher level of detail in IRR-it and PWC-Net-it: the biker's brim, the pant leg, etc. }
\label{fig:4k_1}
\end{figure}

\begin{figure}[!htb]
\begin{tabular}{cc}
\includegraphics[width=.5\linewidth]{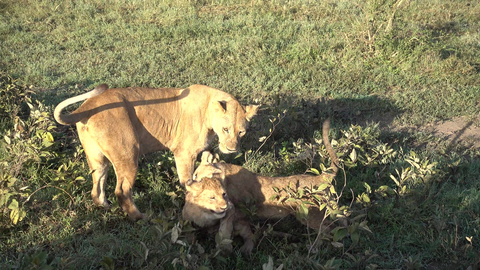}
& \includegraphics[width=.5\linewidth]{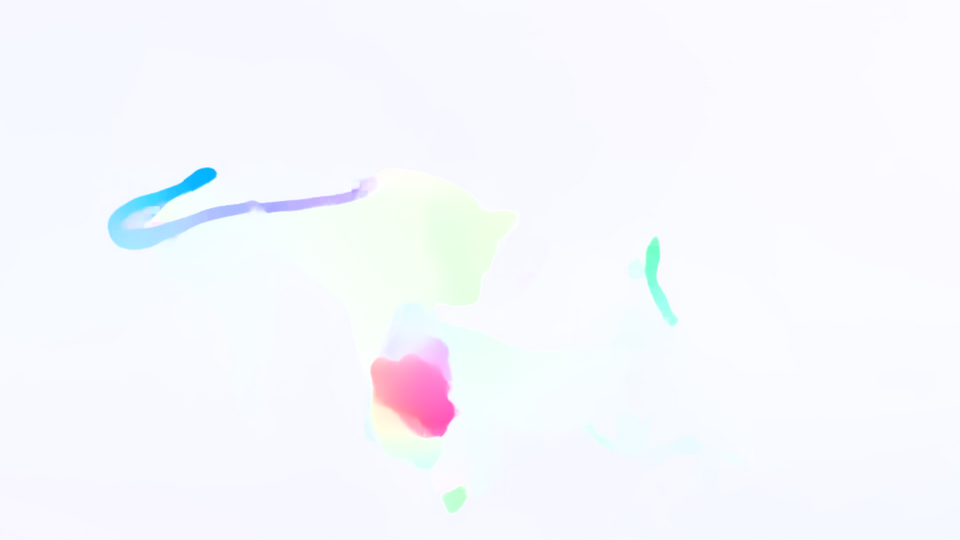} \\
{\scriptsize Frame 0} & {\scriptsize RAFT-it (down-up)} \\ 
\includegraphics[width=.5\linewidth]{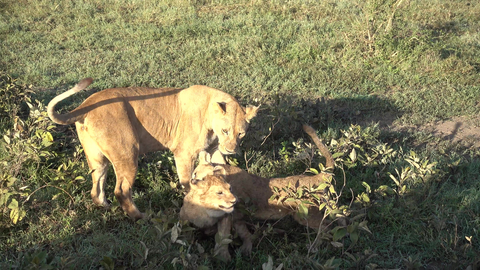}
& {\includegraphics[width=.5\linewidth]{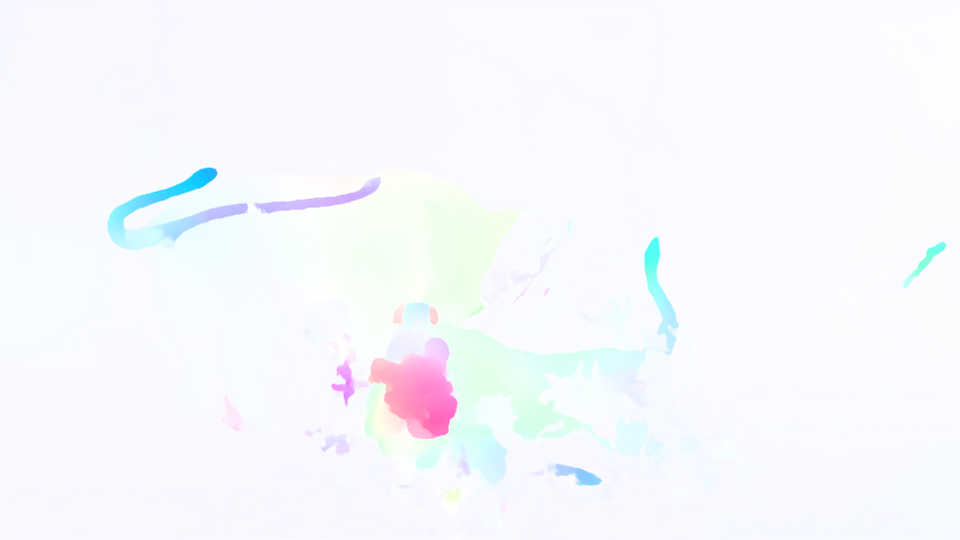}} \\
{\scriptsize Frame 1} & {\scriptsize IRR-it} \\ 
& {\includegraphics[width=.5\linewidth]{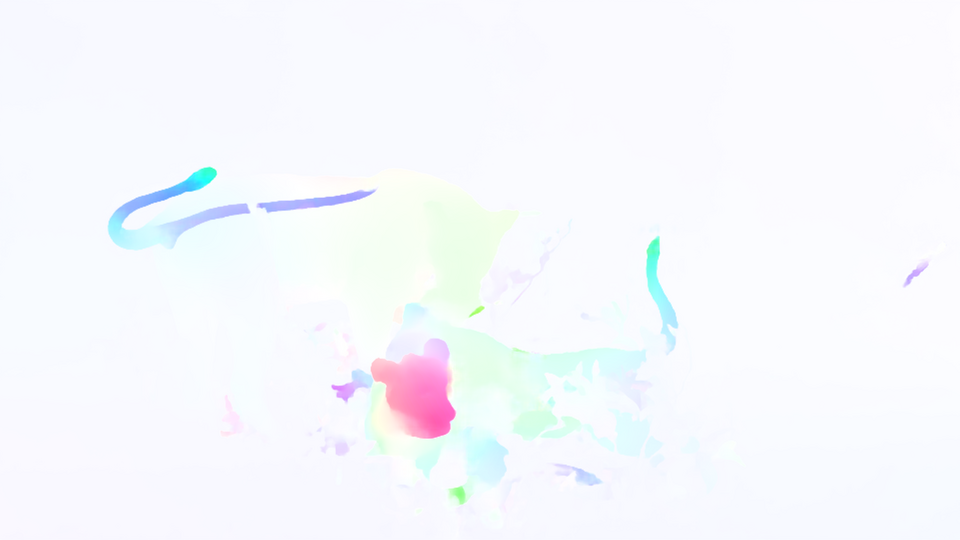}} \\
 & {\scriptsize PWC-Net-it} \\ 
\end{tabular}
\caption{PWC-it, IRR-it, and RAFT-it (down-up) on Davis 4K images. Note that RAFT-it requires the input images to be downsampled due to memory requirements and then the flow upsampled. Note the higher level of detail in IRR-it and PWC-Net-it: plant in front of the front lion. }
\label{fig:4k_6}
\end{figure}

\begin{figure}[!htb]
\begin{tabular}{cc}
\includegraphics[width=.5\linewidth]{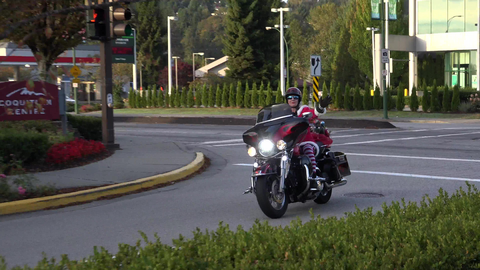}
& \includegraphics[width=.5\linewidth]{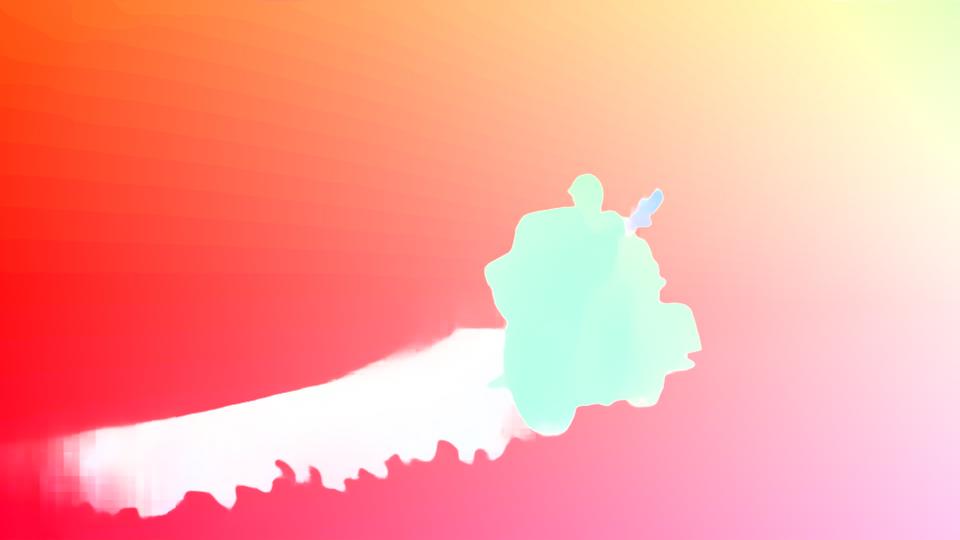} \\
{\scriptsize Frame 0} & {\scriptsize RAFT-it (down-up)} \\ 
\includegraphics[width=.5\linewidth]{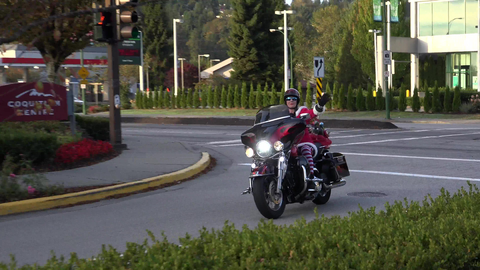}
& {\includegraphics[width=.5\linewidth]{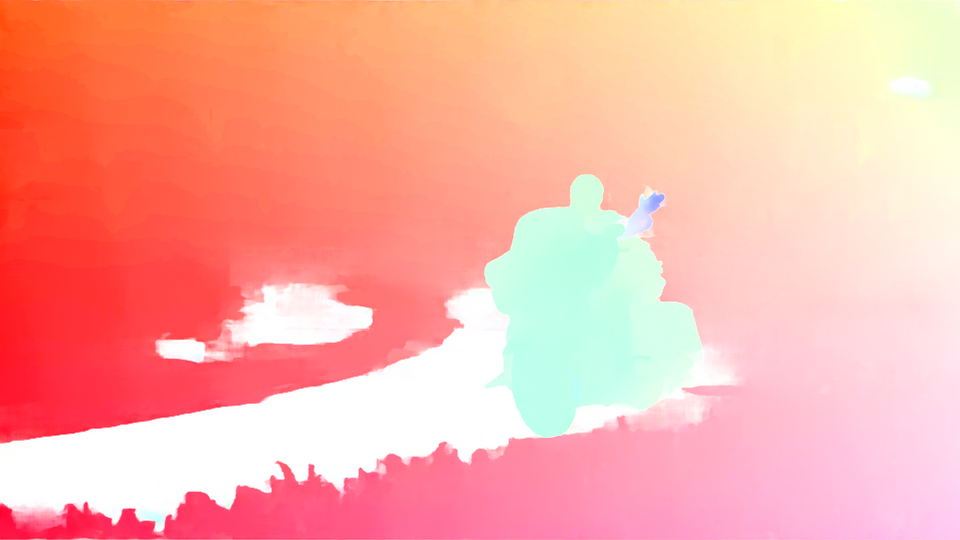}} \\
{\scriptsize Frame 1} & {\scriptsize IRR-it} \\ 
& {\includegraphics[width=.5\linewidth]{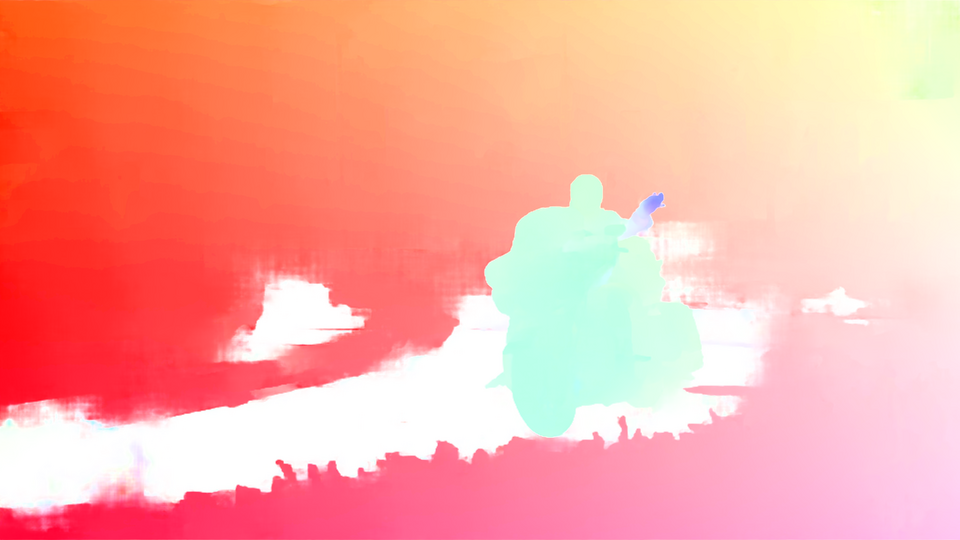}} \\
 & {\scriptsize PWC-Net-it} \\ 
\end{tabular}
\caption{PWC-it, IRR-it, and RAFT-it (down-up) on Davis 4K images. Note that RAFT-it requires the input images to be downsampled due to memory requirements and then the flow upsampled. Note the higher level of detail in IRR-it and PWC-Net-it: the plants in the foreground. }
\label{fig:4k_7}
\end{figure}

In this subsection, we include 4 examples of our improved training models on Davis 4K images: Figures \ref{fig:4k_5}, \ref{fig:4k_1}, \ref{fig:4k_6}, and  \ref{fig:4k_7}. Note that due to memory constraints, RAFT-it requires the input to be downsampled and then the output flow to be upsampled to 4K.

\ignore{
\begin{figure}[!htb]
\begin{tabular}{cc}
\includegraphics[width=.5\linewidth]{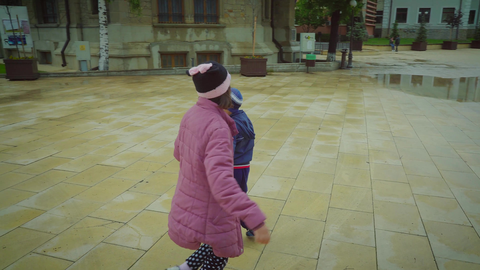}
& \includegraphics[width=.5\linewidth]{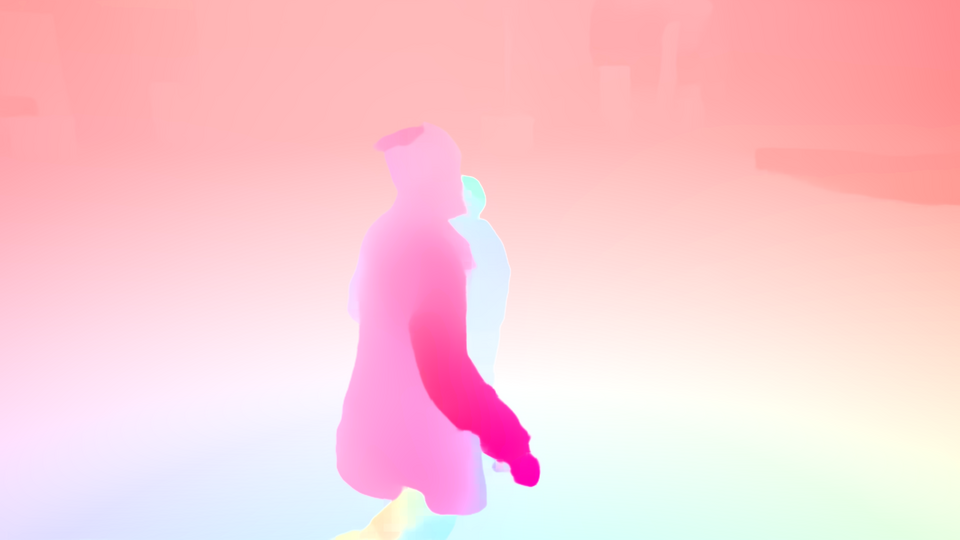} \\
{\scriptsize Frame 0} & {\scriptsize RAFT-it (down-up)} \\ 
\includegraphics[width=.5\linewidth]{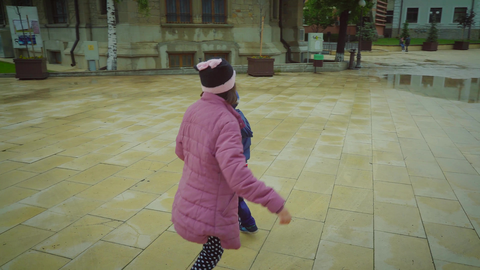}
& {\includegraphics[width=.5\linewidth]{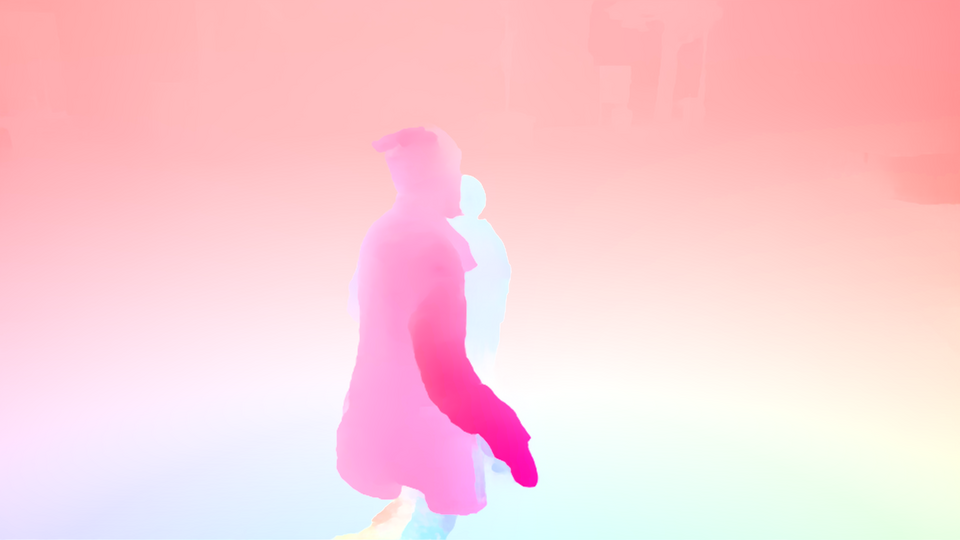}} \\
{\scriptsize Frame 1} & {\scriptsize IRR-it} \\ 
& {\includegraphics[width=.5\linewidth]{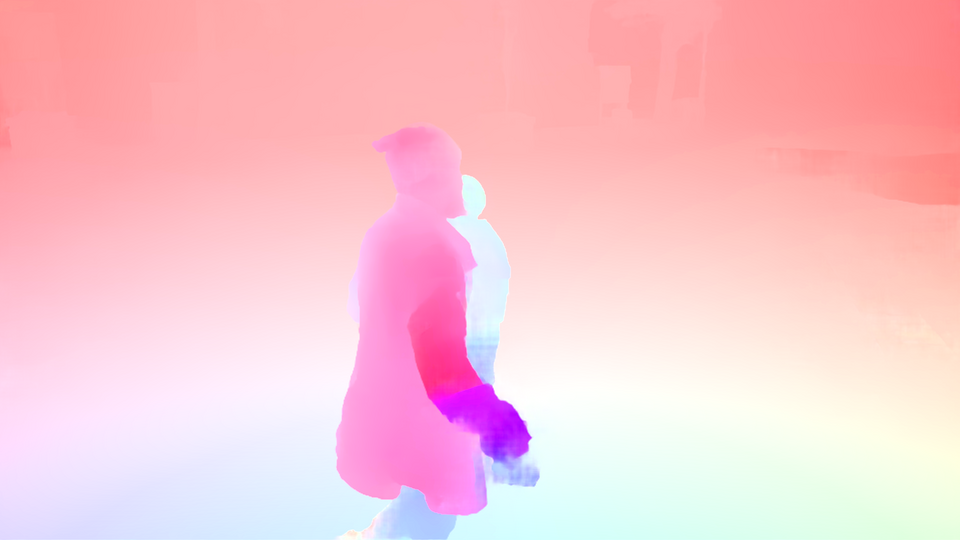}} \\
 & {\scriptsize PWC-Net-it} \\ 
\end{tabular}
\caption{PWC-it, IRR-it, and RAFT-it (down-up) on Davis 4K images. Note that RAFT-it requires the input images to be downsampled due to memory requirements and then the flow upsampled. Note the higher level of detail in IRR-it and PWC-Net-it: the clothes on the person on the left and the person's mouth. }
\label{}
\end{figure}
}

\subsection{Old vs New for PWC and RAFT on Davis 448x864}

\begin{figure}[!htb]
\def\tabcolsep{4pt}
\begin{tabular}{c@{\hskip 2mm}ccc}
{\scriptsize Inputs} & & {\scriptsize PWC-Net} & {\scriptsize RAFT} \\
\noindent\parbox[t]{0.3\linewidth}{\includegraphics[width=\hsize]{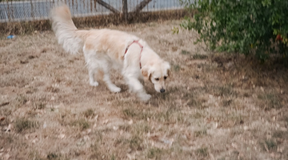}} & \raisebox{4mm}{\rotatebox{90}{\scriptsize Original}} &
\parbox[t]{0.3\linewidth}{\includegraphics[width=\hsize]{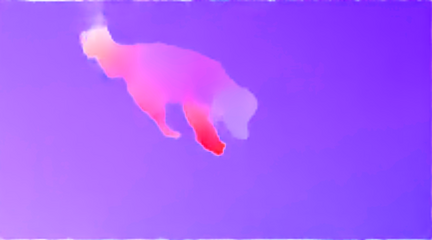}} &
\parbox[t]{0.3\linewidth}{\includegraphics[width=\hsize]{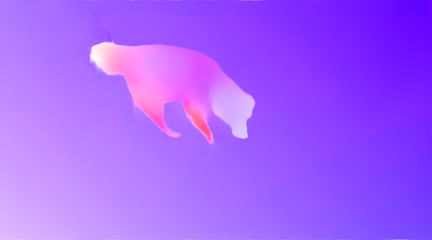}} \\
\noindent\parbox[t]{0.3\linewidth}{\includegraphics[width=\hsize]{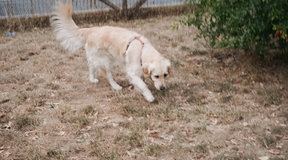}} & \raisebox{6mm}{\rotatebox{90}{{\scriptsize New}}} &
\parbox[t]{0.3\linewidth}{\includegraphics[width=\hsize]{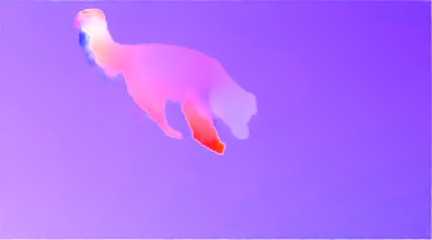}} &
\parbox[t]{0.3\linewidth}{\includegraphics[width=\hsize]{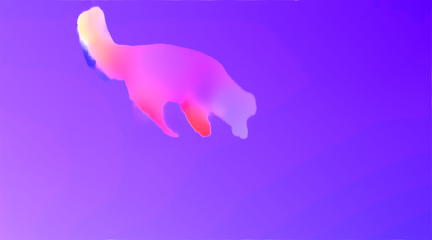}} \\
{\scriptsize Inputs} & & {\scriptsize PWC-Net} & {\scriptsize RAFT} \\
\noindent\parbox[t]{0.3\linewidth}{\includegraphics[width=\hsize]{figures_supp/viz_oldnew_pwcraft_c/davis_scene13_frame10_im0.png}} & \raisebox{4mm}{\rotatebox{90}{\scriptsize Original}} &
\parbox[t]{0.3\linewidth}{\includegraphics[width=\hsize]{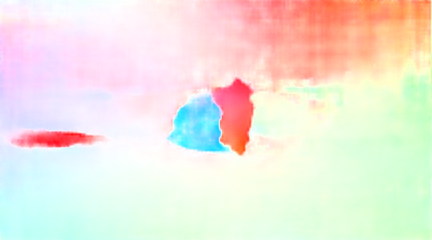}} &
\parbox[t]{0.3\linewidth}{\includegraphics[width=\hsize]{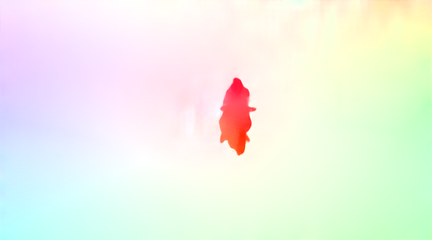}} \\
\noindent\parbox[t]{0.3\linewidth}{\includegraphics[width=\hsize]{figures_supp/viz_oldnew_pwcraft_c/davis_scene13_frame10_im1.png}} & \raisebox{6mm}{\rotatebox{90}{{\scriptsize New}}} &
\parbox[t]{0.3\linewidth}{\includegraphics[width=\hsize]{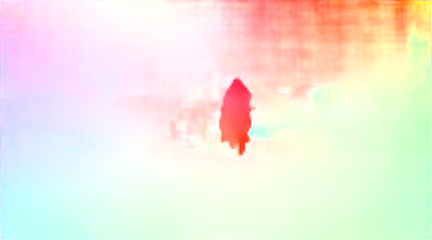}} &
\parbox[t]{0.3\linewidth}{\includegraphics[width=\hsize]{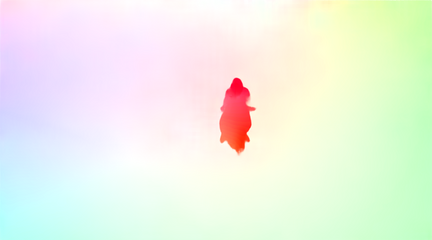}} \\
{\scriptsize Inputs} & & {\scriptsize PWC-Net} & {\scriptsize RAFT} \\
\noindent\parbox[t]{0.3\linewidth}{\includegraphics[width=\hsize]{figures_supp/viz_oldnew_pwcraft_c/davis_scene14_frame10_im0.png}} & \raisebox{4mm}{\rotatebox{90}{\scriptsize Original}} &
\parbox[t]{0.3\linewidth}{\includegraphics[width=\hsize]{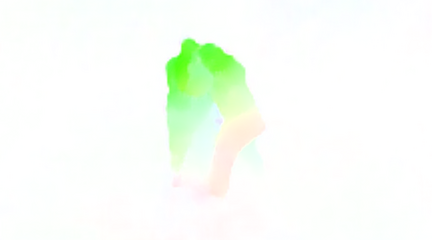}} &
\parbox[t]{0.3\linewidth}{\includegraphics[width=\hsize]{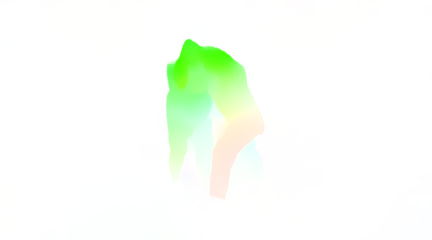}} \\
\noindent\parbox[t]{0.3\linewidth}{\includegraphics[width=\hsize]{figures_supp/viz_oldnew_pwcraft_c/davis_scene14_frame10_im1.png}} & \raisebox{6mm}{\rotatebox{90}{{\scriptsize New}}} &
\parbox[t]{0.3\linewidth}{\includegraphics[width=\hsize]{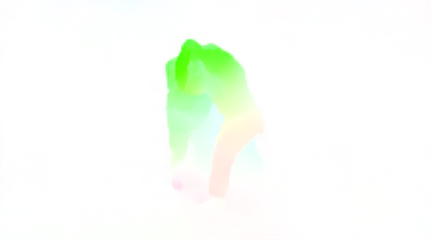}} &
\parbox[t]{0.3\linewidth}{\includegraphics[width=\hsize]{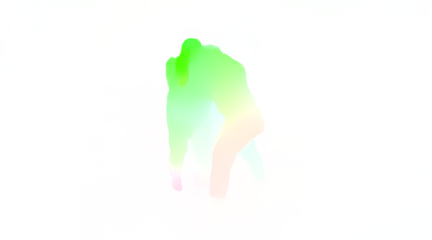}} \\
{\scriptsize Inputs} & & {\scriptsize PWC-Net} & {\scriptsize RAFT} \\
\noindent\parbox[t]{0.3\linewidth}{\includegraphics[width=\hsize]{figures_supp/viz_oldnew_pwcraft_c/davis_scene16_frame10_im0.png}} & \raisebox{4mm}{\rotatebox{90}{\scriptsize Original}} &
\parbox[t]{0.3\linewidth}{\includegraphics[width=\hsize]{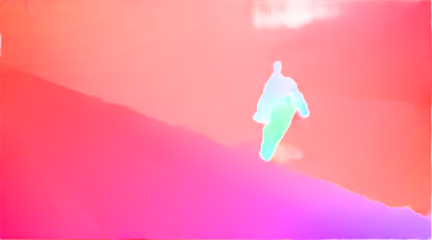}} &
\parbox[t]{0.3\linewidth}{\includegraphics[width=\hsize]{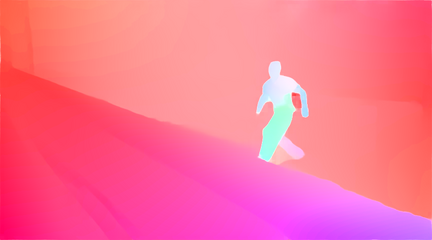}} \\
\noindent\parbox[t]{0.3\linewidth}{\includegraphics[width=\hsize]{figures_supp/viz_oldnew_pwcraft_c/davis_scene16_frame10_im1.png}} & \raisebox{6mm}{\rotatebox{90}{{\scriptsize New}}} &
\parbox[t]{0.3\linewidth}{\includegraphics[width=\hsize]{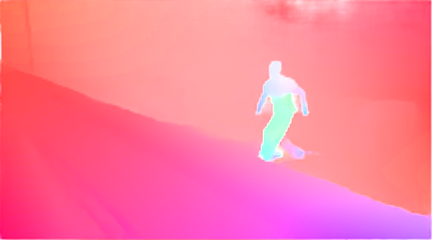}} &
\parbox[t]{0.3\linewidth}{\includegraphics[width=\hsize]{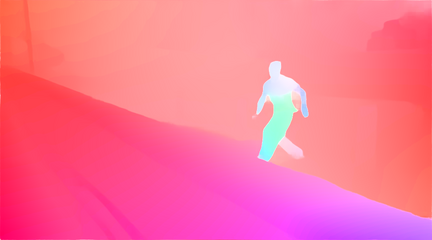}} \\
\end{tabular}
\caption{PWC-orig, RAFT-orig vs PWC-it, RAFT-it on Davis 448x864}
\label{fig:vs_davissmall_1}
\end{figure}

In this subsection, we include 8 examples of two-frame optical flow from PWC-original, RAFT-original, PWC-it, and RAFT-it evaluated on Davis images with resolution 448 by 864: Figures \ref{fig:vs_davissmall_1}.

\subsection{Old vs New for PWC and RAFT on Viper 1080x1920}

\begin{figure}[!htb]
\def\tabcolsep{4pt}
\begin{tabular}{c@{\hskip 2mm}ccc}
{\scriptsize Inputs} & & {\scriptsize PWC-Net} & {\scriptsize RAFT} \\
\noindent\parbox[t]{0.3\linewidth}{\includegraphics[width=\hsize]{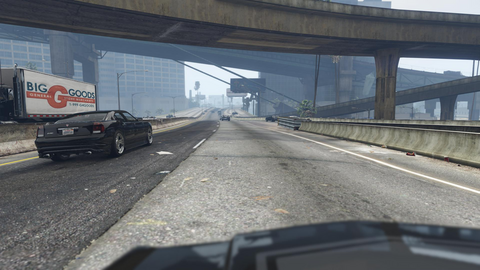}} & \raisebox{4mm}{\rotatebox{90}{\scriptsize Original}} &
\parbox[t]{0.3\linewidth}{\includegraphics[width=\hsize]{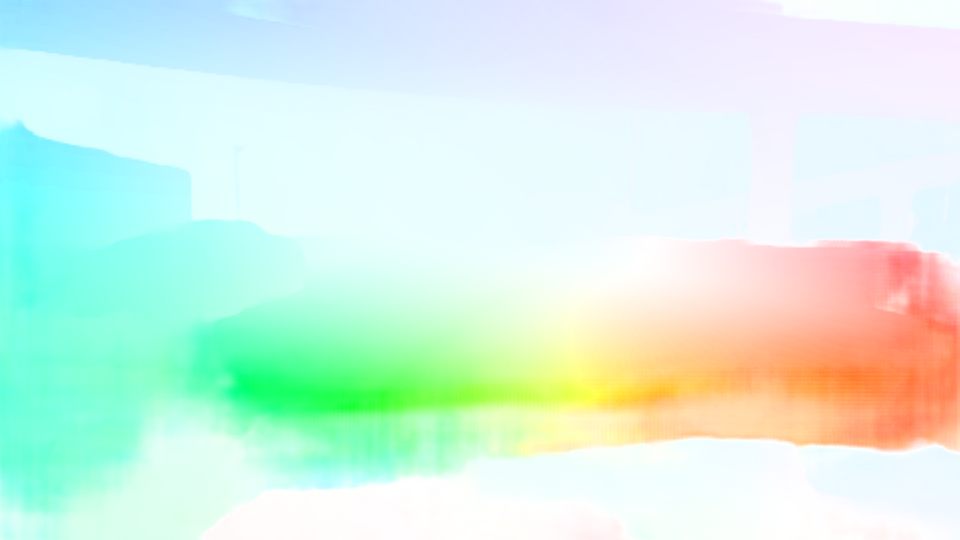}} &
\parbox[t]{0.3\linewidth}{\includegraphics[width=\hsize]{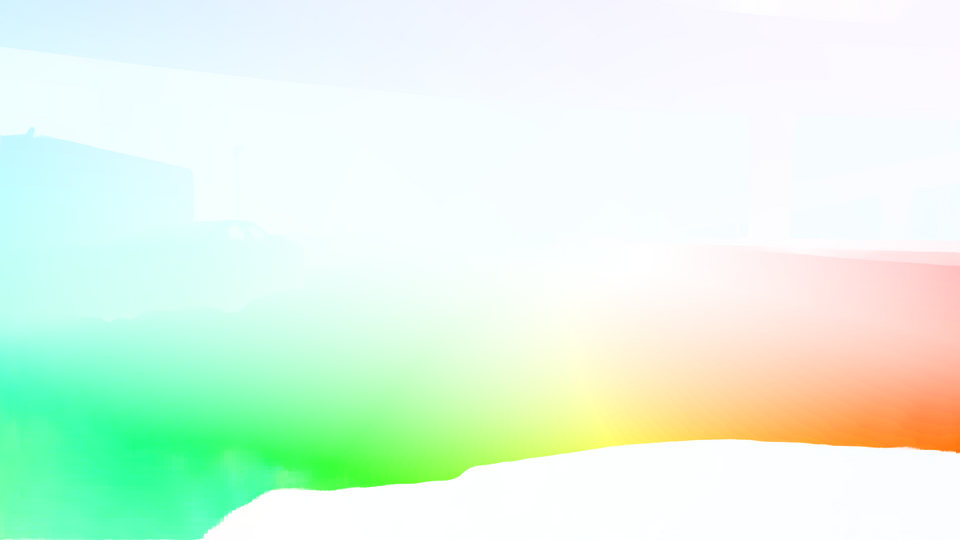}} \\
\noindent\parbox[t]{0.3\linewidth}{\includegraphics[width=\hsize]{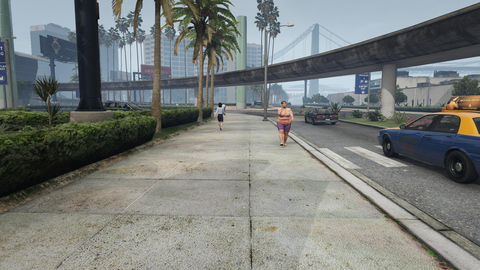}} & \raisebox{6mm}{\rotatebox{90}{{\scriptsize New}}} &
\parbox[t]{0.3\linewidth}{\includegraphics[width=\hsize]{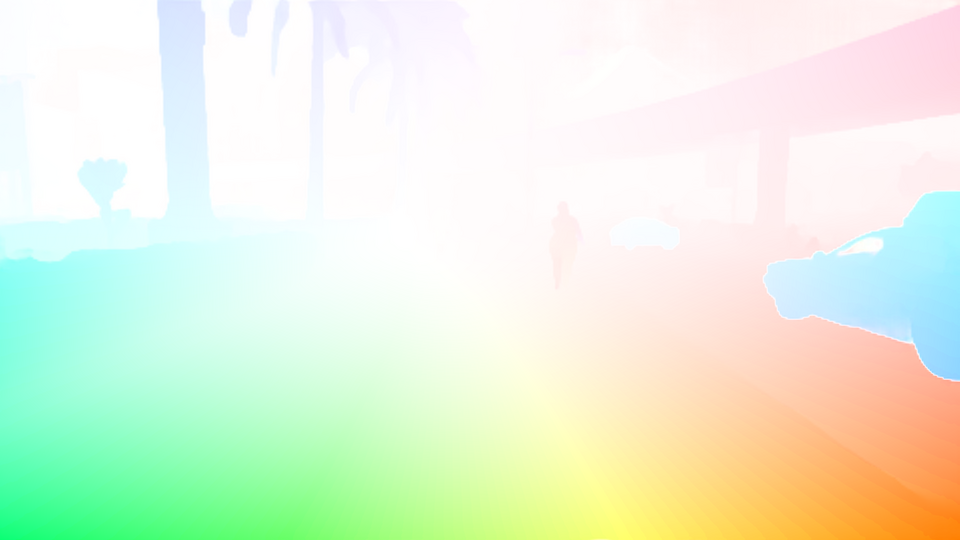}} &
\parbox[t]{0.3\linewidth}{\includegraphics[width=\hsize]{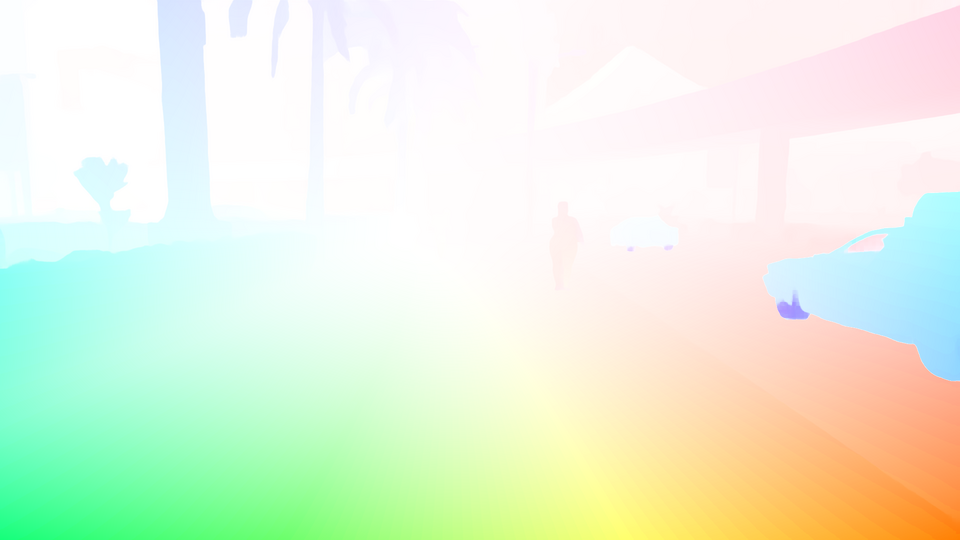}} \\
{\scriptsize Inputs} & & {\scriptsize PWC-Net} & {\scriptsize RAFT} \\
\noindent\parbox[t]{0.3\linewidth}{\includegraphics[width=\hsize]{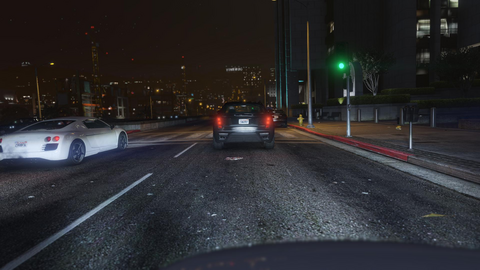}} & \raisebox{4mm}{\rotatebox{90}{\scriptsize Original}} &
\parbox[t]{0.3\linewidth}{\includegraphics[width=\hsize]{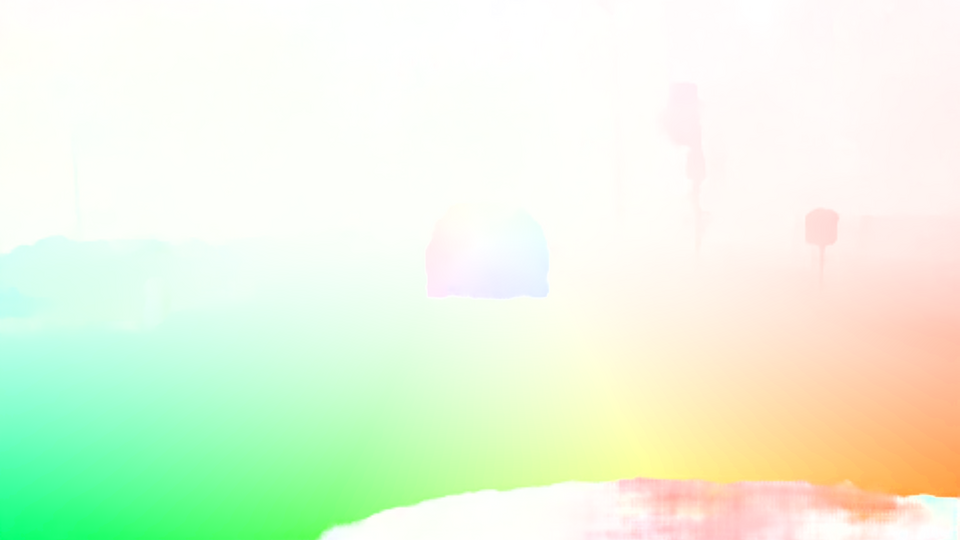}} &
\parbox[t]{0.3\linewidth}{\includegraphics[width=\hsize]{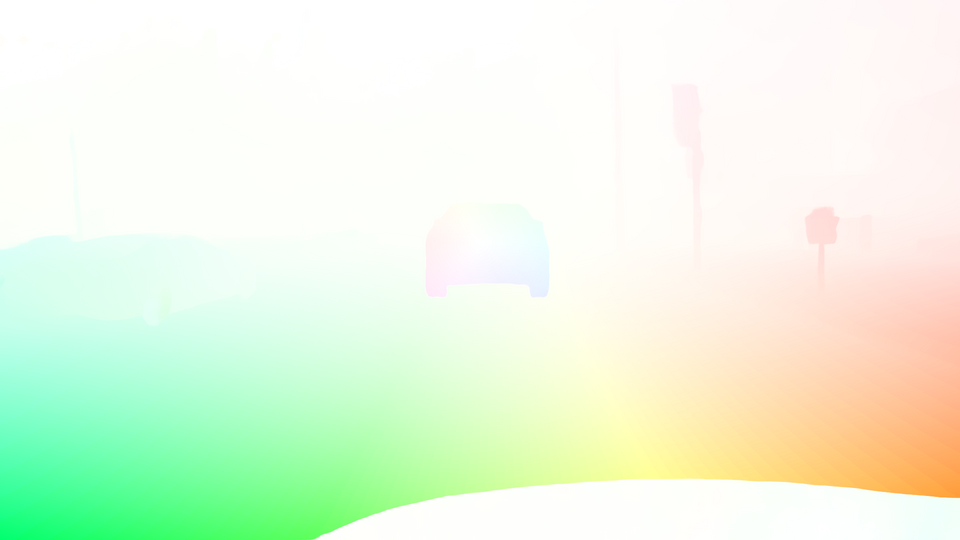}} \\
\noindent\parbox[t]{0.3\linewidth}{\includegraphics[width=\hsize]{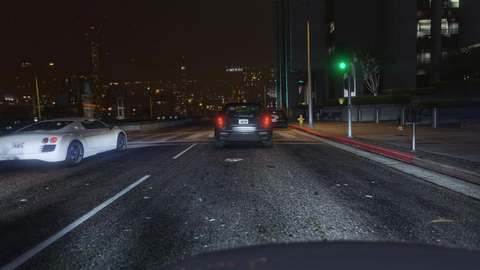}} & \raisebox{6mm}{\rotatebox{90}{{\scriptsize New}}} &
\parbox[t]{0.3\linewidth}{\includegraphics[width=\hsize]{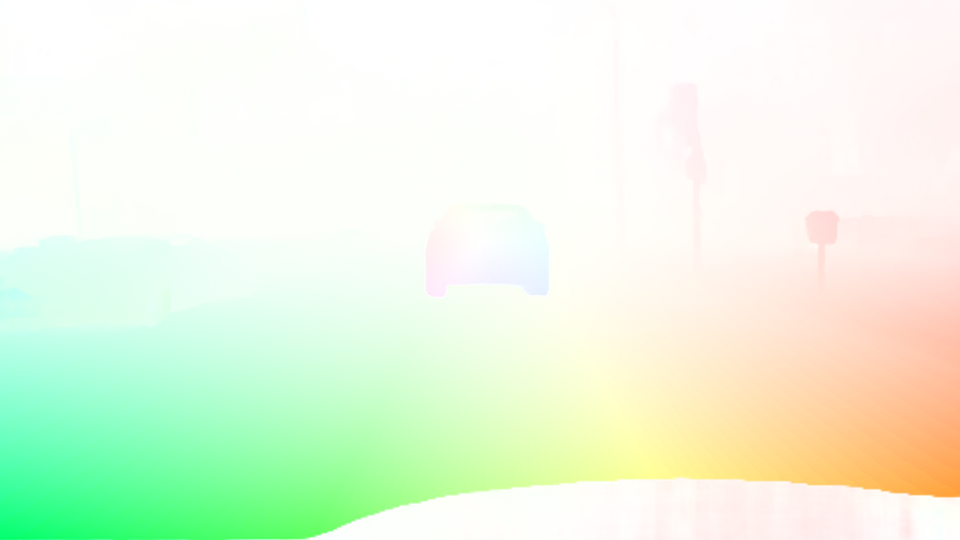}} &
\parbox[t]{0.3\linewidth}{\includegraphics[width=\hsize]{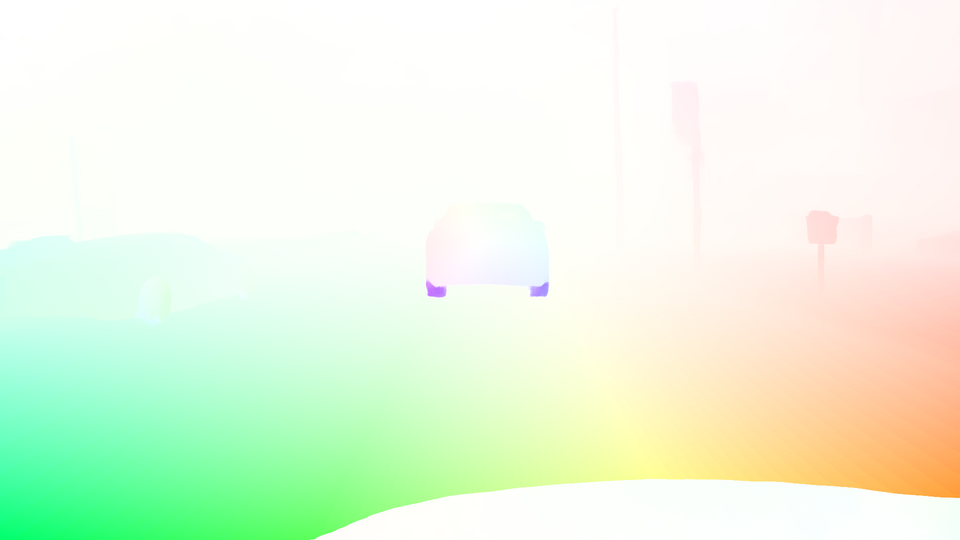}} \\
{\scriptsize Inputs} & & {\scriptsize PWC-Net} & {\scriptsize RAFT} \\
\noindent\parbox[t]{0.3\linewidth}{\includegraphics[width=\hsize]{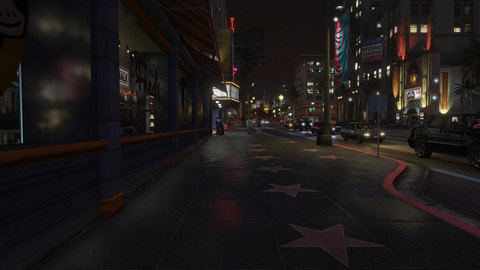}} & \raisebox{4mm}{\rotatebox{90}{\scriptsize Original}} &
\parbox[t]{0.3\linewidth}{\includegraphics[width=\hsize]{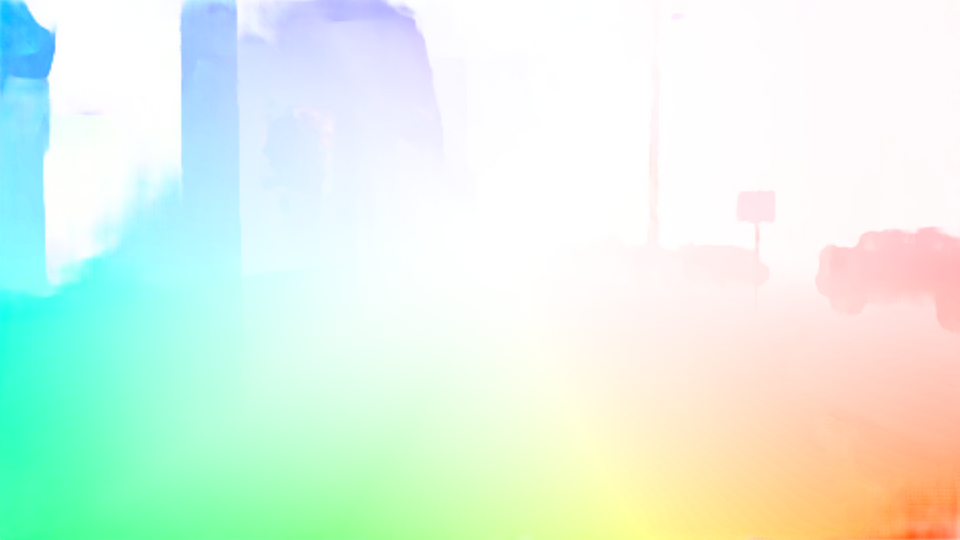}} &
\parbox[t]{0.3\linewidth}{\includegraphics[width=\hsize]{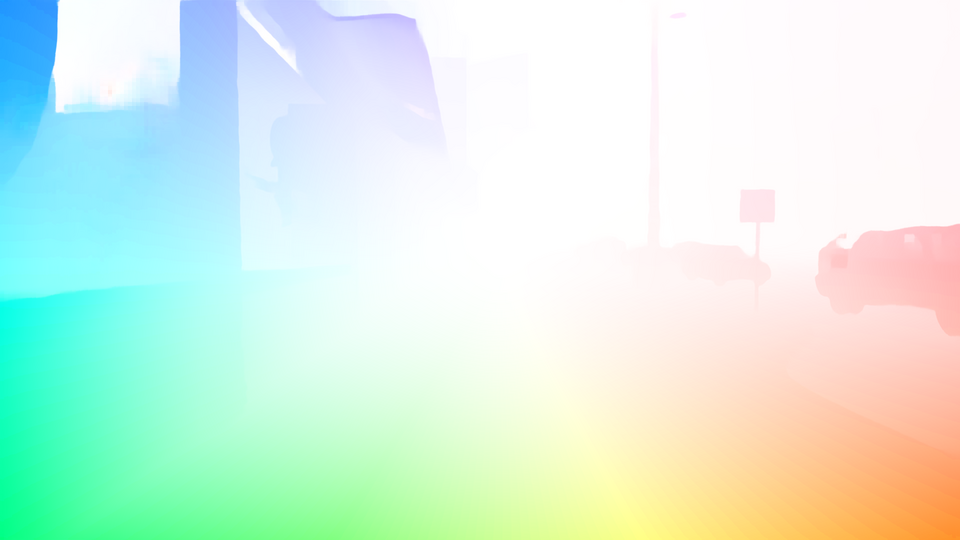}} \\
\noindent\parbox[t]{0.3\linewidth}{\includegraphics[width=\hsize]{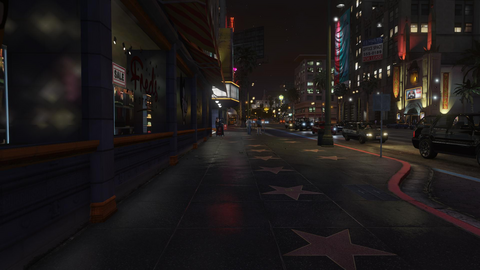}} & \raisebox{6mm}{\rotatebox{90}{{\scriptsize New}}} &
\parbox[t]{0.3\linewidth}{\includegraphics[width=\hsize]{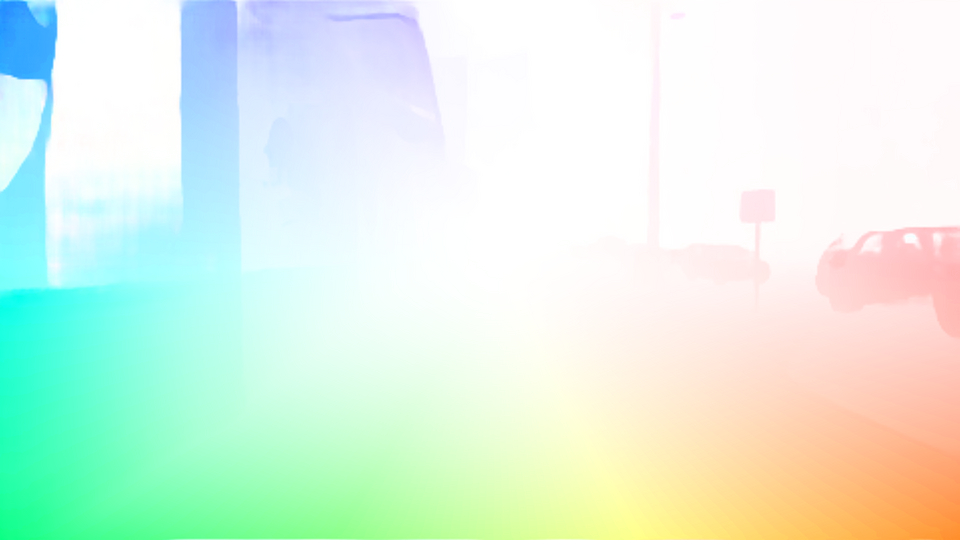}} &
\parbox[t]{0.3\linewidth}{\includegraphics[width=\hsize]{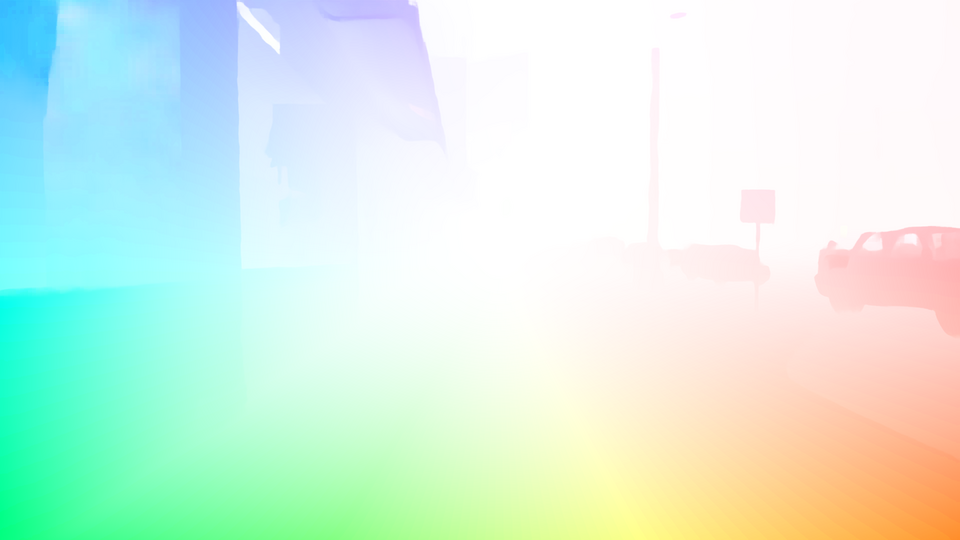}} \\
\end{tabular}
\caption{PWC-orig, RAFT-orig vs PWC-it, RAFT-it on Viper 1080x1920}
\label{fig:vs_vipersmall_1}
\end{figure}

In this subsection, we include 3 examples of two-frame optical flow from PWC-original, RAFT-original, PWC-it, and RAFT-it evaluated on Viper validation images with resolution 1080 by 1920: Figures \ref{fig:vs_vipersmall_1}.

\end{document}